# GENERALIZED OPTIMAL MATCHING METHODS FOR CAUSAL INFERENCE

By Nathan Kallus

*Cornell University*

We develop an encompassing framework for matching, covariate balancing, and doubly-robust methods for causal inference from observational data called generalized optimal matching (GOM). The framework is given by generalizing a new functional-analytical formulation of optimal matching, giving rise to the class of GOM methods, for which we provide a single unified theory to analyze tractability, consistency, and efficiency. Many commonly used existing methods are included in GOM and, using their GOM interpretation, can be extended to optimally and automatically trade off balance for variance and outperform their standard counterparts. As a subclass, GOM gives rise to kernel optimal matching (KOM), which, as supported by new theoretical and empirical results, is notable for combining many of the positive properties of other methods in one. KOM, which is solved as a linearly-constrained convex-quadratic optimization problem, inherits both the interpretability and model-free consistency of matching but can also achieve the $\sqrt{n}$-consistency of well-specified regression and the efficiency and robustness of doubly robust methods. In settings of limited overlap, KOM enables a very transparent method for interval estimation for partial identification and robust coverage. We demonstrate these benefits in examples with both synthetic and real data.

**1. Introduction.** In causal inference, matching is the pursuit of comparability between samples that differ in systematic ways due to selection (often self-selection) by way of subsampling or re-weighting the samples [1]. Optimal matching [2], wherein each treated unit[1] is matched to one or more control units to minimize some objective (such as sum) in the list of within-match pairwise distances so to optimize comparability[2] and implemented in the popular $R$ package `optmatch`, is arguably one of the most commonly used methods for causal inference on treatment effects, whether used as an estimator or as a preprocessing step before regression analysis [4].

Since the introduction of optimal matching, a variety of other methods for matching on covariates have been developed, including coarsened exact matching [5], genetic matching [6], combining optimal matching with near-fine balance on one stratification [7], and using integer programming to match sample mean vectors for simultaneous near-fine balance on multiple stratifications [3]. These have largely been developed independently and ad hoc given distinct motivations and definitions for what is balance and how to improve it relative to no matching, optimally or non-optimally. There are also a variety of other methods for causal inference on treatment effects such as regression analysis [8], propensity score matching and weighting [9], and doubly robust methods that combine the latter two [10].

---

[1] In the context of estimating the effect on the treated.

[2] The term "optimal matching" has also been used in other contexts such as optimal near-fine balance [3], but the most common usage by far, which we follow here, refers to matching made on one-to-one, one-to-many, or many-to-many basis using network flow and bipartite approaches, such as optimal bipartite (one-to-one) matching with weights equal to covariate vector distances.

In this paper, we develop an encompassing framework and theory for matching and weighting methods and related methods for causal inference that reveal the connections and motivations behind these various existing methods and, moreover, give rise to new and improved ones. We begin by providing a functional analytical characterization of optimal matching as a weighting method that minimizes worst-case conditional mean squared error given the observed data and assumptions on (a) the space of feasible conditional expectation functions, (b) the space of feasible weights, and (c) the magnitude of residual variance. By generalizing the lattermost, we develop a new optimal matching method that correctly and automatically accounts for the balance-variance trade-off inherent in matching and by doing so can reduce effect estimation error. By generalizing all three and using functional analysis and modern optimization, we develop a new class of generalized optimal matching (GOM) methods that construct matched samples or distributions of the units to eliminate imbalances. It turns out that many existing methods are included in GOM, including nearest-neighbor matching, one-to-one matching, optimal caliper matching, coarsened exact matching, various near-fine balance approaches, and linear regression adjustment. Moreover, using the lens of GOM many of these too are extended to new methods that judiciously and automatically trade off balance for variance and that outperform their standard matching counterparts. We provide theory on both tractability and consistency that applies generally to GOM methods.

Finally, as a subclass of GOM, we develop kernel optimal matching (KOM), which is particularly notable for combining the interpretability and potential use as preprocessing of matching methods [4], the non-parametric nature and model-free consistency of optimal matching [2, 11], the $\sqrt{n}$-consistency of well-specified regression-based estimators [8], the efficiency [12] and robustness [10] of augmented inverse propensity weight estimators, the careful selection of matched sample size of monotonic imbalance bounding methods [13], and the model-selection flexibility of Gaussian-process regression [14]. We show that KOM can be interpreted as Bayesian efficient in a certain sense, that it is computationally tractable, and that it is consistent. We discuss how to tune the hyperparameters of KOM and demonstrate the efficacy of doing so. KOM allows for a transparent way to bound any irreducible biases due to a lack of overlap between control and treated populations, which leads to robust interval estimates that can partially identify effects in a highly interpretable manner. We develop the augmented kernel weighted estimator and establish robustness and efficiency guarantees for it related to those of the augmented inverse propensity weighted estimator. Furthermore, we establish similar guarantees for KOM used as a preprocessing step before linear regression, rigorously establishing that it reduces model dependence and yields efficient estimation under a well-specified model without ceding model-free consistency under misspecification. We end with a discussion on relevant connections to and non-linear generalizations of equal percent bias reduction. We study the practical usefulness of KOM by applying the new methods developed to a semi-simulated case study using real data and find that KOM offers significant benefits in efficiency but also in robustness to practical issues like limited overlap and lack of model specification.

**2. Re-interpreting Optimal Matching.** In this section we present the first building blocks toward generalizing optimal matching. We set up the causal estimation problem and provide a bias-variance decomposition of error. Through a new functional analytical lens on optimal matching, we uncover it as a specific case of finding weights that minimize worst-case error, but only under zero residual variance of outcomes given covariates. Our first generalization is to consider non-zero residual variance, giving rise to a balance-variance efficient version of optimal matching and to a method that automatically chooses the exchange between balance and variance.

2.1. *Setting.* The observed data consists of $n$ independent and identically distributed (iid) observations $\{(X_i, T_i, Y_i) : i = 1, \ldots, n\}$ of the variables $(X, T, Y)$, where $X \in \mathcal{X}$ denotes baseline covariates, $T \in \{0, 1\}$ treatment assignment, and $Y \in \mathbb{R}$ outcome. The space $\mathcal{X}$ is general; assumptions about it will be specified as necessary. For $t = 0, 1$, we let $\mathcal{T}_t = \{i : T_i = t\}$ and $n_t = |\mathcal{T}_t|$. We also let $T_{1:n} = (T_1, \ldots, T_n)$ and $X_{1:n} = (X_1, \ldots, X_n)$ denote all the observed treatment assignments and baseline covariates, respectively. Using Neyman-Rubin potential outcome notation [15, Ch. 2], we let $Y_i(0), Y_i(1)$ be the real-valued potential outcomes for unit $i$ and assume the stable unit treatment value assumption [16] holds. We let $Y_i = Y_i(T_i)$, capturing consistency and non-interference. We define

$$f_0(x) = \mathbb{E}\left[Y(0) \mid X = x\right], \quad \epsilon_i = Y_i(0) - f_0(X_i), \quad \sigma_i^2 = \mathrm{Var}\left(Y_i(0) \mid X_i\right).$$

We consider estimating the *sample average treatment effect on the treated*:

$$\text{SATT} = \tfrac{1}{n_1} \sum_{i \in \mathcal{T}_1} (Y_i(1) - Y_i(0)) = \overline{Y}_{\mathcal{T}_1}(1) - \overline{Y}_{\mathcal{T}_1}(0),$$

where $\overline{Y}_{\mathcal{T}_t}(s) = \tfrac{1}{n_t} \sum_{i \in \mathcal{T}_t} Y(s)$ is the average outcome of treatment $s$ in the $t$-treated sample. As $\overline{Y}_{\mathcal{T}_1}(1)$ is observed, we consider estimators of the form

$$\hat{\tau} = \overline{Y}_{\mathcal{T}_1}(1) - \hat{\overline{Y}}_{\mathcal{T}_1}(0)$$

for some choice of $\hat{\overline{Y}}_{\mathcal{T}_1}(0)$. We will focus on *weighting* estimators $\hat{\overline{Y}}_{\mathcal{T}_1}(0) = \sum_{i \in \mathcal{T}_0} W_i Y_i$ given weights $W \in \mathbb{R}^{\mathcal{T}_0}$. Moreover, we restrict to *honest* weights that only depend on the observed $X_{1:n}, T_{1:n}$ and not on observed outcome data, that is, $W = W(X_{1:n}, T_{1:n})$. The resulting estimator has the form

$$(2.1) \qquad \hat{\tau}_W = \tfrac{1}{n_1} \sum_{i \in \mathcal{T}_1} Y_i - \sum_{i \in \mathcal{T}_0} W_i Y_i.$$

An alternative weighting estimator, which we call the augmented weighting (AW) estimator, can be derived as a generalization of the doubly-robust augmented inverse propensity weighting (AIPW) estimator [10, 17, 18]:

$$(2.2) \qquad \hat{\tau}_{W, \hat{f}_0} = \tfrac{1}{n_1} \sum_{i \in \mathcal{T}_1} (Y_i - \hat{f}_0(X_i)) - \sum_{i \in \mathcal{T}_0} W_i (Y_i - \hat{f}_0(X_i)),$$

where $\hat{f}_0(x)$ is a regression estimator for $f_0(x)$. The standard AIPW for SATT would be given by $W_{\hat{p}, i} = \frac{\hat{p}(X_i)}{n_1(1 - \hat{p}(X_i))}$ where $\hat{p}(x)$ is a binary regression estimator for $\mathbb{P}(T = 1 \mid X = x)$.

Given a data set, we measure the risk of a weighting estimator as its conditional mean squared error (CMSE), conditioned on all the observed data upon which the weights depend as honest weights:

$$\text{CMSE}(\hat{\tau}) = \mathbb{E}\left[ (\hat{\tau} - \text{SATT})^2 \mid X_{1:n}, T_{1:n} \right].$$

When choosing weights $W$, one may restrict to a certain space of allowable weights $\mathcal{W}$. Throughout, we consider only permutation symmetric sets, satisfying $P\mathcal{W} = \mathcal{W}$ for all permutation matrices $P \in \mathbb{R}^{\mathcal{T}_0 \times \mathcal{T}_0}$. For example, $\mathcal{W}^{\text{general}} = \mathbb{R}^{\mathcal{T}_0}$ allows *all* weights; $\mathcal{W}^{\text{simplex}} = \left\{ W \geq 0 : \sum_{i \in \mathcal{T}_0} W_i = 1 \right\}$ restricts to weights that give a probability measure, preserving the unit of analysis and ensuring no extrapolation in estimating $\overline{Y}_{\mathcal{T}_1}(0)$; $\mathcal{W}^{b\text{-simplex}} = \mathcal{W}^{\text{simplex}} \cap [0, b]^{\mathcal{T}_0}$ further bounds how much weight we can put on a single unit; $\mathcal{W}^{n_0'\text{-multisubset}} = \mathcal{W}^{\text{simplex}} \cap \{0, 1/n_0', 2/n_0', \dots\}^{\mathcal{T}_0}$ limits us to integer-multiple weights that exactly correspond to sub-sampling a multisubset of the control units of cardinality $n_0'$; $\mathcal{W}^{n_0'\text{-subset}} = \mathcal{W}^{\text{simplex}} \cap \{0, 1/n_0'\}^{\mathcal{T}_0}$ corresponds to sub-sampling a usual subset of cardinality $n_0'$; and $\mathcal{W}^{\text{multisubsets}} =$ $\cup_{n_0'=1}^{n_0} \mathcal{W}^{n_0'\text{-multisubset}}$ and $\mathcal{W}^{\text{subsets}} = \cup_{n_0'=1}^{n_0} \mathcal{W}^{n_0'\text{-subset}}$ correspond to sub-sampling any multisubset or subset, respectively. We have the inclusions:

$$(2.3) \qquad \mathcal{W}^{\text{subsets}} \subseteq \mathcal{W}^{\text{multisubsets}} \subseteq \mathcal{W}^{\text{simplex}} \subseteq \mathcal{W}^{\text{general}}.$$

A standing assumption in this paper is that of *weak mean-ignorability*, a weaker form of ignorability [9].

ASSUMPTION 1. *For each $t = 0, 1$, conditioned on $X$, $Y(t)$ is mean-independent of $T$, that is, $\mathbb{E}\left[Y(t) \mid T, X\right] = \mathbb{E}\left[Y(t) \mid X\right]$.*

A second assumption, which we will discuss relaxing in the context of partial identification, is *overlap*.

ASSUMPTION 2. $\mathbb{P}\left(T = 0 \mid X\right)$ *is bounded away from 0.*

2.2. *Decomposing the Conditional Mean Squared Error.* In this section we decompose the CMSE of estimators of the form in eq. (2.1) into a bias term and a variance term. Let us define

$$B(W; f) = \frac{1}{n_1} \sum_{i \in \mathcal{T}_1} f(X_i) - \sum_{i \in \mathcal{T}_0} W_i f(X_i)$$
$$V^2(W; \sigma_{1:n}^2) = \sum_{i \in \mathcal{T}_0} W_i^2 \sigma_i^2 + \frac{1}{n_1^2} \sum_{i \in \mathcal{T}_1} \sigma_i^2$$
$$E^2(W; f_0, \sigma_{1:n}^2) = B^2(W; f_0) + V^2(W; \sigma_{1:n}^2)$$

THEOREM 1. *Under Asn. 1,[3]*

$$\mathbb{E}\left[\hat{\tau}_W - \text{SATT} \mid X_{1:n}, T_{1:n}\right] = B(W; f_0), \quad \text{CMSE}(\hat{\tau}_W) = E^2(W; f_0, \sigma_{1:n}^2).$$

The above provides a decomposition of the risk of $\hat{\tau}_W$ into a (conditional) bias term and a (conditional) variance term, which must be balanced to minimize overall risk. The first term, $B(W; f_0)$, is exactly the conditional bias of $\hat{\tau}_W$. We refer to $V^2(W; f_0)$ as the variance term of the error.[4] More generally, if the units are not independent, the proof makes clear that Thm. 1 holds with the variance term $(W, e_{n_1}/n_1)^T \Sigma (W, e_{n_1}/n_1)$ where $e_{n_1}$ is the vector of all ones of length $n_1$ and $\Sigma$ is the conditional covariance matrix.

An analogous result holds for the AW estimator when $\hat{f}_0$ is fitted to an independent sample. (Cross-fold fitting is discussed briefly in Sec. 3.6.)

---

[3]Note the use of $f_0$ as the true conditional expectation function of $Y(0)$ and $f$ as a generic function-valued variable in the space of all functions $\mathcal{X} \to \mathbb{R}$.

[4]The conditional variance of $\hat{\tau}_W$ actually differs from $V^2(W; \sigma_{1:n}^2)$ by exactly $\frac{1}{n_1^2} \sum_{i \in \mathcal{T}_1} \left(\text{Var}\left(Y_i(1) \mid X_i\right) - \text{Var}\left(Y_i(0) \mid X_i\right)\right)$, which accounts for the conditional variance of SATT and its covariance with $\hat{\tau}_W$. Note this difference is constant in $W$ and so it does not matter whether the estimand we consider is SATT or CSATT $= \mathbb{E}\left[\text{SATT} \mid X_{1:n}, T_{1:n}\right]$.

COROLLARY 2.    *Under Asn. 1, if $\hat{f}_0 \perp\!\!\!\perp Y_{1:n} \mid X_{1:n}, T_{1:n}$ then*

$$\mathbb{E}\left[\hat{\tau}_{W,f_0} - \text{SATT} \mid X_{1:n}, T_{1:n}\right] = B(W; f_0 - \hat{f}_0),$$
$$\text{CMSE}(\hat{\tau}_{W,f_0}) = E^2(W; f_0 - \hat{f}_0, \sigma^2_{1:n})$$

2.3. *Re-interpreting Optimal Matching.*    In this section, we provide an interpretation of optimal matching as minimizing worst-case CMSE. We consider two forms of optimal matching: nearest neighbor matching (NNM) and optimal one-to-one matching (1:1M). In both, each treated unit is matched to one control unit to minimize the sum of distances between matches as measured by a given extended pseudo-metric $\delta(X_i, X_j)$.[5] NNM allows for *replacement* of control units whereas 1:1M does not. In the end, the weight $W_i$ assigned to a control unit $i \in \mathcal{T}_0$ is equal to $1/n_1$ times the number of times it has been matched. So, under 1:1M, $W_i$ is capped at $1/n_1$ and the result is equivalent to constructing a subset of cardinality $n_1$, where all $n_0 - n_1$ unmatched control units have been pruned away. Under NNM, the result is equivalent to a *multi*-subset of the control sample of cardinality $n_1$.

Next, consider an alternative perspective. We seek weights $W$ that depend only on data $X_{1:n}, T_{1:n}$ and that minimize the resulting CMSE. The CMSE depends on unknowns: $f_0$ and $\sigma^2_{1:n}$. In order to get a handle on the CMSE, we make assumptions about these unknowns. First, we assume that $X_i$ is completely predictive of $Y_i(0)$ so that $\sigma^2_i = 0$. Second, we assume that $f_0$ is a Lipschitz continuous function with respect to $\delta$. That is,

$$\exists \gamma \geq 0 : \|f_0\|_{\text{Lip}(\delta)} \leq \gamma \quad \text{where} \quad \|f\|_{\text{Lip}(\delta)} := \sup_{x \neq x'} \frac{f(x) - f(x')}{\delta(x, x')} \leq \gamma.$$

Assuming nothing else, we may seek $W$ to minimize the worst-case CMSE. If we limit ourselves to simplex weights $\mathcal{W} = \mathcal{W}^{\text{simplex}}$, the next theorem shows that this is precisely equivalent to optimal matching.

THEOREM 3.    *Fix a pseudo-metric $\delta : \mathcal{X} \times \mathcal{X} \to \mathbb{R}_+$. Then, for any $\gamma > 0$, NNM and 1:1M are equivalent to*

$$(2.4) \qquad W \in \text{argmin}_{W \in \mathcal{W}} \sup_{\|f\|_{Lip(\delta)} \leq \gamma} \left\{ E^2(W; f, \mathbf{0}) = B^2(W; f) \right\},$$

*where $\mathcal{W} = \mathcal{W}^{simplex}$ for NNM and $\mathcal{W} = \mathcal{W}^{1/n_1\text{-}simplex}$ for 1:1M.*

Therefore, optimal matching is indeed optimal in a minimax CMSE sense, given the assumptions and restrictions made. That is, the above theorem

---

[5]Compared to a metric, an extended pseudo-metric may also assign zero or infinity distance to distinct elements. Any proper metric is also an extended pseudo-metric.

Fig 1: The Balance-Variance Trade-off in Optimal Matching

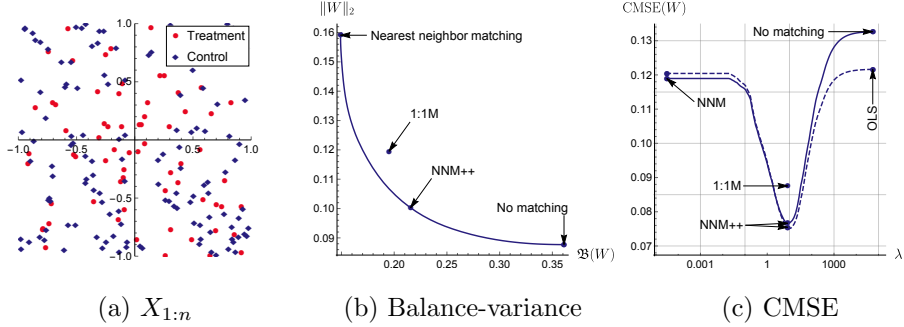

(a) $X_{1:n}$      (b) Balance-variance      (c) CMSE

relates optimal matching – an *a priori* choice of re-weighting based on data $X_{1:n}, T_{1:n}$ – to the *a posteriori* error of estimating a causal treatment effect and says that this choice minimizes the worst-case error over all $\gamma$-Lipschitz functions. Surprisingly, we need not restrict to selecting each control unit an integer multiple number of times – optimal matching, which results in a subset or multisubset of the control sample, minimizes this error among all continuous weights in the (bounded) simplex.

It is well-known that optimal matching can be formulated as a linear optimization problem, specifically a minimum-cost network flow problem [2]. Indeed, optimal matching minimizes the Wasserstein metric (also known as the earth mover's distance) between the matched subsamples. The Wasserstein metric is an example of integral probability metrics (IPM) [19], which are distance metrics between measures that take the form

$$d_{\mathcal{F}}(\mu, \nu) = \sup_{f \in \mathcal{F}} \int f d(\mu - \nu),$$

given some class $\mathcal{F}$. The above reinterpretation arises from linear optimization duality and is closely related to the Rubinstein-Kantorovich theorem that establishes the dual forms of the Wasserstein metric [20].

This reinterpretation of optimal matching involved three critical choices: a restriction on the conditional expectation $f_0$, a restriction on the space weights $\mathcal{W}$, and a restriction on the magnitude of residual variance $\sigma_{1:n}^2$. In this paper, we consider different such choices that lead to methods that *generalize* optimal matching.

**3. Generalizing Optimal Matching.** In this section we consider generalizing the restrictions and assumptions that made optimal matching equivalent to minimizing worst-case error. Doing so recovers other common match-ing methods and other causal estimation methods, as well as give rise to new matching methods such as KOM.

3.1. *Generalizing Balance.* Balance between the control and treatment samples can be understood as the extent to which they are comparable. In estimating SATT, we want the samples to be comparable on their values of $f_0$ so that the bias due to the systematic differences between the samples is minimal. When we re-weight the control sample by $W$, the absolute discrepancy in values of $f_0$ is precisely $|B(W; f_0)|$. As seen in the preceding section, by minimizing this quantity over all possible realizations of Lipschitz functions, optimal matching is seeking the best possible balance over this class of functions. We now generalize this functional restriction, leading to more general balance metrics called bias-dual-norm balance metrics [21].

Since the bias depends on $f_0$ but we do not know $f_0$, we consider guarding against any reasonable realization of $f_0$. Bias is linear in $f_0$, *i.e.*, $B(W; \alpha f + \alpha' f') = \alpha B(W; f) + \alpha' B(W; f')$. So, we must limit the "size" of $f_0$. In particular, we consider the bias relative to some extended magnitude $\|f\| \in [0, \infty]$ of $f_0$ that is absolutely homogeneous, *i.e.*, $\|\alpha f\| = |\alpha| \|f\|$ (where $|\alpha| \infty = \infty$), and satisfies the triangle inequality (where $\infty \leq \infty$). This allows us to generalize the notion of balance for optimal matching. Letting $0/0 = 0$, we redefine imbalance more generally as

$$\mathfrak{B}(W; \|\cdot\|) = \sup_{f:\mathcal{X}\to\mathbb{R}} B(W; f)/\|f\| = \sup_{\|f\|\leq 1} B(W; f),$$

where the last equality is due to the homogeneity of $B(W; \cdot)$ and $\|\cdot\|$. This too is in fact an IPM between the treated sample and re-weighted control sample. It is not clear, however, whether it is well-defined.

To ensure that it is well-defined, we restrict our attention only to certain magnitude functions. We require that $B(W; \cdot)$ is bounded with respect to $\|\cdot\|$, *i.e.*, $\forall W \in \mathcal{W}$ $\exists M_W > 0 : B(W; f) \leq M_W \|f\|$.[6] Then $\{f : \|f\| < \infty\}$ is a semi-normed[7] vector space and $B(W; \cdot)$ is a well-defined, continuous, linear operator on the Banach completion of the quotient space $\{f : \|f\| < \infty\} / \{f : \|f\| = 0\}$, *i.e.*, $B(W; \cdot)$ is in its dual space. (See [22, 23] for Banach spaces.) In particular, $\mathfrak{B}(W; \|\cdot\|)$ is precisely the *dual norm of the bias as an operator on conditional expectation functions*, which is necessarily finite and well-defined for any $W \in \mathcal{W}$:

$$\mathfrak{B}(W; \|\cdot\|) = \|B(W; \cdot)\|_* < \infty.$$

---

[6]This is necessary: were $B(W; \cdot)$ not bounded for some $W \in \mathcal{W}$ then for any $M > 0$ we would have some $f$ with $B(W; f) > M\|f\|$ so that indeed $\mathfrak{B}(W; \|\cdot\|) = \infty$ is not defined.

[7]Compared to a norm, a semi-norm may assign zero magnitude to non-zeros.

DEFINITION 1. Given $\mathcal{W}$ and $\| \cdot \| : [\mathcal{X} \to \mathbb{R}] \to [0, \infty]$ such that $\| \cdot \|$ is absolutely homogeneous and satisfies the triangle inequality and $\forall W \in \mathcal{W} \ \exists M_W > 0 : |B(W; f)| \leq M_W \|f\|$,[8] $\mathfrak{B}(W; \| \cdot \|)$ is called a *bias-dual-norm (BDN) imbalance metric* on $\mathcal{W}$.

3.2. *The Balance-Variance Trade-off.* Even if the control and treatment samples are made completely comparable, there is inherent error to the estimation of outcomes in each sample. Just a few controls may provide the best matches, and hence the least bias. But, if $\sigma_i^2$ are nonzero, then averaging the outcome in these few units has higher variance than averaging more units by, *e.g.*, finding a few more but perhaps less good matches. In one extreme, if $\sigma_i^2$ is zero, there is no added variance and we best use the best matches (*e.g.*, reuse controls with replacement). In the other extreme, we conceive of $\sigma_i^2$ being so large, that we do not care about the bias due to imbalance and we would prefer to do *no matching* on the samples so to minimize variance (assuming homoskedasticity) and estimate SATT as the simple mean difference of the raw treated and control samples.

This trade off between bias and variance is well understood in matching [24, 25, 26, 13]. The most common approach to this trade off in optimal matching is to disallow replacement (1:1M instead of NNM) and to increase the number of matches (1:$k$M instead of 1:1M). But given an explicit understanding of balance as bounding bias, these are only heuristic and need not be on the efficient frontier of achievable balance and variance.

Per Thm. 3, NNM is given by optimizing only balance and ignoring variance. Some approaches like 1:1M seek to alleviate this by forcing a more even distribution of weights. However, per Thm. 1, given an imbalance metric, the best way to trade off balance and variance for minimal error is by directly regularizing imbalance by the sum of squared weights. This suggests that the right way to trade off balance and variance in optimal matching is to consider eq. (2.4) with $\sigma_{1:n}^2 \neq \mathbf{0}$. Plugging $\sigma_i^2 = \lambda\gamma^2$ for $\lambda \geq 0$ into eq. (2.4), we refer to the result as Balance-Variance Efficient Nearest Neighbor Matching (BVENNM). We will revisit BVENNM in Sec. 5 and develop NNM++, which automatically selects $\lambda$ using cross-validation. For now, we explore the balance-variance trade-off in an example.

In general, moving beyond optimal matching toward GOM, Thm. 1 provides an explicit form of the total estimation risk in terms of these competing objectives and suggests that the best choice lies somewhere in between focus-ing solely on balance or solely on variance, where balance can be understood more broadly than sum of matched-pair distances.

---

[8]Note that this condition is relative to $\mathcal{W}$. For example, for $\| \cdot \| = \| \cdot \|_{\mathrm{Lip}(\delta)}$, the condition does hold for $\mathcal{W} = \mathcal{W}^{\mathrm{simplex}}$ but does *not* hold for $\mathcal{W} = \mathcal{W}^{\mathrm{general}}$. So the balance metric in optimal matching is *not* valid for general non-simplex weights.

EXAMPLE 1. Let $X \sim \text{Unif}[-1, 1]^2$, $\mathbb{P}(T = 1 \mid X) = 0.95/(1 + \frac{3}{\sqrt{2}} \|X\|_2)$. Fix a draw of $X_{1:n}, T_{1:n}$ with $n = 200$. We plot the resulting draw, which has $n_0 = 130, n_1 = 70$, in Fig. 1a. For a range of $\lambda$ we compute the resulting BVENNM weights using the Mahalanobis distance $\delta(x, x') = (x - x')\hat{\Sigma}_0^{-1}(x - x')$ where $\hat{\Sigma}_0$ is the sample covariance of $X \mid T = 0$. We plot the resulting space of achievable balance and variance in Fig. 1b. In one extreme ($\lambda = 0$) we have NNM and in the other ($\lambda = \infty$) we have no matching. Since 1:1M minimizes the same balance criterion (sum of pair distances), we can plot the balance and variance it achieves on the same axes. As intended, 1:1M achieves a trade off between the two extremes, but it is not actually on the efficient frontier since it does not trade these off in the *optimal* way. Next, let $Y(0) \mid X \sim \mathcal{N}(\|X\|_2^2 - e^T X/2, \sqrt{3})$. In Fig. 1c, varying $\lambda$, we plot the resulting CMSE of $\hat{\tau}_W$ (solid) and $\hat{\tau}_{W, \hat{f}_0}$ (dashed) for $\hat{f}_0$ given by ordinary least squares (OLS). Since OLS has in-sample residuals summing to zero, $\hat{\tau}_{W, \hat{f}_0}$ for $\lambda = \infty$ corresponds to simple OLS regression adjustment. We see that tuning $\lambda$ correctly can amount to a significant improvement in CMSE.

3.3. *Generalized Optimal Matching Methods.* Optimal matching minimized the worst-case squared error given certain restrictions on $f_0$, $\sigma_{1:n}^2$, and $\mathcal{W}$. Generalizing these restrictions, we can consider a whole range of generalized optimal matching methods that minimize CMSE by trading off variance to new, generalized notions of balance.

DEFINITION 2. Given $\mathcal{W}, \|\cdot\|$ satisfying the assumptions of Def. 1 and $\lambda \in [0, \infty]$, the generalized optimal matching method GOM($\mathcal{W}, \|\cdot\|, \lambda$) is given by the weights $W$ that solve[9]

$$(3.1) \qquad \min_{W \in \mathcal{W}} \left\{ \mathfrak{E}^2(W; \|\cdot\|, \lambda) := \mathfrak{B}^2(W; \|\cdot\|) + \lambda \|W\|_2^2 \right\}.$$

We let $\mathfrak{E}_{\min}^2(\mathcal{W}, \|\cdot\|, \lambda)$ denote the value of this minimum.

More generally, if we have knowledge of heteroskedasticity or even if the units are not independent, we would take the second term to be $W^T \Lambda W$ for some positive semi-definite $\Lambda$. In this paper, we focus only on $\Lambda = \lambda I$ for the sake of simplicity. Our consistency results, nonetheless, will apply under heteroskedasticity even when we use a single $\lambda$.

---

[9]If $\lambda = \infty$, then $W$ minimizes the first term over the minimizers of the second term.

Thm. 3 established that NNM is equivalent to GOM($\mathcal{W}^{\text{simplex}}, \|\cdot\|_{\text{Lip}(\delta)}, 0$) and 1:1M is equivalent to GOM($\mathcal{W}^{1/n_1\text{-simplex}}, \|\cdot\|_{\text{Lip}(\delta)}, 0$). BVENNM is given by GOM($\mathcal{W}^{\text{simplex}}, \|\cdot\|_{\text{Lip}(\delta)}, \lambda$). Similarly, for any $\|\cdot\|$, no matching is given by GOM($\mathcal{W}, \|\cdot\|, \infty$) for any $\mathcal{W} \ni (1/n_0, \ldots, 1/n_0)$, examples of which include $\mathcal{W}^{\text{simplex}}$, $\mathcal{W}^{\text{multisubsets}}$, and $\mathcal{W}^{\text{subsets}}$.

It follows by Thm. 1 that GOM leads to a bound on the CMSE. Define

$$\|[f]\| = \inf_{g:B(W;g)=0 \ \forall W \in \mathcal{W}} \|f + g\|,$$

which acts on the quotient space that eliminates degrees of freedom that are irrelevant to $B(W; f)$. For example, when $\mathcal{W} \subseteq \mathcal{W}^{\text{simplex}}$, this includes all constant shifts. Note $\|[f]\|$ is always smaller than $\|f\|$.

COROLLARY 4.   *Suppose $\sigma^2 \geq \sigma_t^2$ and $\gamma \geq \|[f_0]\|$. Let $\lambda = \sigma^2/\gamma^2$ and let $W$ be given by* GOM($\mathcal{W}, \|\cdot\|, \lambda$). *Then*

$$\text{CMSE}(\hat{\tau}_W) \leq \gamma^2(\mathfrak{E}^2_{\min}(\mathcal{W}, \|\cdot\|, \lambda) + \lambda/n_1).$$

*And, if $\hat{f}_0 \perp\!\!\!\perp Y_{1:n} \mid X_{1:n}, T_{1:n}$ and $\gamma \geq \|[f_0 - \hat{f}_0]\|$ , then*

$$\text{CMSE}(\hat{\tau}_{W,\hat{f}_0}) \leq \gamma^2(\mathfrak{E}^2_{\min}(\mathcal{W}, \|\cdot\|, \lambda) + \lambda/n_1).$$

For subset-based matching, the balance-variance efficient frontier given by varying $\lambda$ is given by solely-balance-optimizing fixed-sized subsets.

THEOREM 5.   *Given $\|\cdot\|$ and $\lambda \in [0, \infty]$, there exists $n(\lambda) \in \{1, \ldots, n_0\}$ such that* GOM($\mathcal{W}^{\text{subsets}}, \|\cdot\|, \lambda$) *is equivalent to* GOM($\mathcal{W}^{n(\lambda)\text{-subset}}, \|\cdot\|, 0$).

In particular, to compute GOM($\mathcal{W}^{\text{subsets}}, \|\cdot\|, \lambda$) we may search over GOM($\mathcal{W}^{n_0'\text{-subset}}, \|\cdot\|, 0$) for $n_0' \in \{1, \ldots, n_0\}$ and pick the one that minimizes $\mathfrak{E}(W; \|\cdot\|, \lambda)$. Note that the converse is not true: there may be some cardinalities that are *not* on the efficient frontier of balance-variance efficient subsets. An example of this will be seen in Ex. 6. We also have the following relationship between optimal fixed-cardinality multisubsets and subsets:

THEOREM 6.   *Given $\|\cdot\|$, $\lambda \in [0, \infty]$ and $n_0' \in \{1, \ldots, n_0\}$, the following are equivalent:* GOM($\mathcal{W}^{n_0'\text{-mutlisubset}}, \|\cdot\|, \infty$), GOM($\mathcal{W}^{n_0'\text{-subset}}, \|\cdot\|, 0$), *and* GOM($\mathcal{W}^{n_0'\text{-subset}}, \|\cdot\|, \lambda$).

3.4. *Tractability.*   GOM is given by an optimization problem, which begs the question of when is it computationally tractable. We can first establish that the objective is always convex.

THEOREM 7. *Given any $\mathcal{W}, \|\cdot\|$ satisfying the assumptions of Def. 1, $\mathfrak{E}^2(W; \|\cdot\|, \lambda)$ is convex in $W$.*

This means that if $\mathcal{W} = \mathcal{W}^{\text{simplex}}$ then problem (3.1) is convex. Indeed, we can show that we can solve it in polynomial time.

THEOREM 8. *Given an evaluation oracle for $\mathfrak{B}(W; \|\cdot\|)$, we can solve problem (3.1) for $\mathcal{W} = \mathcal{W}^{\text{simplex}}$ up to $\epsilon$ precision in time and oracle calls polynomial in $n, \log(1/\epsilon)$.*

In all cases we consider, $\mathfrak{B}(W; \|\cdot\|)$ will be easy to evaluate. Moreover, in all cases we consider with $\mathcal{W} = \mathcal{W}^{\text{simplex}}$, we will in fact be able to formulate problem (3.1) as a linearly-constrained convex-quadratic optimization problem, which are not only polynomially-time solvable but also easily solved in practice using off-the-shelf solvers like Gurobi (www.gurobi.com), which we use in all numerics in this paper to solve such problems in tens to hundreds of milliseconds on a personal laptop computer. This includes the case of kernel optimal matching, which we introduce in Sec. 4.

If $\mathcal{W} = \mathcal{W}^{n_0'\text{-subset}}$ then, by Thm. 6, problem (3.1) is equivalent to a convex-objective binary optimization problem:

$$(3.2) \qquad \min_{U \in \{0,1\}^{\mathcal{T}_0} : \sum_{i \in \mathcal{T}_0} U_i = n_0'} \mathfrak{B}(U/n_0'; \|\cdot\|).$$

Unlike simplex weights, this problem is *not* polynomial-time solvable.

If $\mathcal{W} = \mathcal{W}^{\text{subsets}}$ then Thm. 5 shows that problem (3.1) is equivalent to searching over the solutions $U_{n_0'}$ to problem (3.2) for $n_0' \in \{1, \ldots, n_0\}$ and picking the one with minimal $\mathfrak{B}(U_{n_0'}/n_0'; \|\cdot\|) + \lambda/n_0'$.

In all cases we consider in this paper with $\mathcal{W} = \mathcal{W}^{n_0'\text{-subset}}$ or $\mathcal{W} = \mathcal{W}^{\text{subsets}}$, we will be able to formulate problem (3.1) as, respectively, a single or a series of either binary quadratic or mixed-integer-linear optimization problem(s). These problems, generally hard in the sense of being NP-hard, can be solved for many practical sizes of $n$ also by Gurobi. In fact, we solve these problems too in our numerical examples.

3.5. *Existing Matching Methods as GOM.* A surprising fact is that many matching methods commonly used in practice – not just NNM and 1:1M – are also GOM. These include optimal-caliper matching (OCM), which GOM with respect to an averaged Lipschitz norm; coarsened exact matching (CEM) [5], which is GOM with respect to the $L_\infty$ norm on piece-wise linear functions; methods that use mean matching, near-fine balance, and combinations thereof with pair matching [3, 27, 28, 29, 7] are also GOM with norms given by parametric spaces and their direct sum with Lipschitz spaces. Like NNM and 1:1M, many of these are GOM with $\lambda = 0$. By automatically selecting $\lambda$ using hyperparameter estimation, we can develop extensions of these methods, such as NNM++ and CEM++, that automatically and optimally trade off balance and variance and reduce overall estimation error. Finally, regression adjustment methods are also GOM, revealing a close connection to matching but also a nuanced but important difference in the handling of extrapolation. For the sake of a more fluid presentation we defer the full presentation of these results, which are summarized in Table 1, to Sec. 5. Instead, we focus first on a new, unified analysis of GOM and the development of KOM as a special class.

TABLE 1

| Method | GOM with | | See |
|---|---|---|---|
| | $\| \cdot \| =$ | $\mathcal{W} =$ | |
| 1:1M | $\| \cdot \|_{\mathrm{Lip}(\delta)}$ | $\mathcal{W}^{1/n_1\text{-simplex}}$ | Thm. 1 |
| NNM | $\| \cdot \|_{\mathrm{Lip}(\delta)}$ | $\mathcal{W}^{\mathrm{simplex}}$ | Thm. 1, Sec. 5.1 |
| Optimal caliper matching | $\| \cdot \|_{\partial(\hat{\mu}_n, \delta)}$ | $\mathcal{W}^{1/n_1\text{-simplex}}$ | Thm. 17, Sec. 5.2 |
| Coarsened exact matching | $\| \cdot \|_{L_\infty(C)}$ | $\mathcal{W}^{\mathrm{simplex}}$ | Thm. 18, Sec. 5.3 |
| Mean-matching and fine balance [27, 28, 29] | $\| \cdot \|_{2-\mathrm{lin}}$ | $\mathcal{W}^{\mathrm{subsets}}$ | Thm. 19, Sec. 5.4 |
| Combined pair- and mean-matching [7, 3] | $\| \cdot \|_{\mathrm{Lip}(\delta)} \oplus_\rho \| \cdot \|_{2-\mathrm{lin}}$ | $\mathcal{W}^{\mathrm{subsets}}$ | Thm. 19, Sec. 5.4 |
| Regression adjustment | $\| \cdot \|_{2-\mathrm{lin}}$ | $\mathcal{W}^{\mathrm{general}}$ | Thms. 21, 22, Sec. 5.6 |

3.6. *Consistency.* In this section we characterize conditions for GOM to lead to consistent estimation. The conditions include "correct specification" by requiring that $\|[f_0]\| < \infty$. For a sequence of random variable $Z_n$ and positive numbers $a_n$, we write $Z_n = o_p(a_n)$ to mean $|Z_n|/a_n$ converges to 0 in probability, *i.e.*, $\forall \epsilon \; \mathbb{P}(|Z_n| \geq a_n \epsilon) \to 0$, and $Z_n = O_p(a_n)$ to mean that $|Z_n|/a_n$ is stochastically bounded, *i.e.*, $\forall \epsilon \; \exists M : \mathbb{P}(|Z_n| > Ma_n) < \epsilon$. Clearly, if $b_n = o(a_n)$ then $Z_n = O_p(b_n)$ implies $Z_n = o_p(a_n)$.

We need the following technical condition on $\| \cdot \|$ for consistency. All magnitudes $\| \cdot \|$ that we consider in this paper satisfy this condition.

DEFINITION 3. $\| \cdot \|$ is *B-convex* if there is $N \in \mathbb{N}, \eta < N$ such that for any $\|g_1\|, \ldots, \|g_N\| \leq 1$ there is a choice of signs so that $\|\pm g_1 \pm \cdots \pm g_N\| \leq \eta$.

THEOREM 9. *Suppose Asns. 1 and 2 hold and that*

(i) *for each $n$, $W$ is given by* $\mathrm{GOM}(\mathcal{W}, \| \cdot \|, \lambda_n)$,

(ii) $\mathcal{W}, \| \cdot \|$ *satisfy the conditions of Def. 1,*

(iii) $\lambda_n \in [\underline{\lambda}, \bar{\lambda}] \subset (0, \infty)$,

*(iv)* $\mathcal{W}^{subsets} \subseteq \mathcal{W}$,
*(v)* $\| \cdot \|$ *is B-convex,*
*(vi)* $\mathbb{E}[\sup_{\|f\| \le 1} (f(X_1) - f(X_2))^2 \mid T_1 = 1, T_2 = 1] < \infty$,
*(vii)* $\mathrm{Var}(Y(0) \mid X)$ *is almost surely bounded, and*
*(viii)* $\|[f_0]\| < \infty$.

*Then,* $\hat{\tau}_W - \mathrm{SATT} = o_p(1)$.

Condition (iv) is satisfied for subset, mutlisubset, and simplex matching. Condition (viii) requires correct specification of the outcome model. For example, for near-fine balance, expected potential outcomes have to be *additive* in the factors (*i.e.*, linear). We will relax this in the case of KOM and prove model-free consistency. The result can be extended to the AW estimator:

THEOREM 10. *Suppose all assumptions of Thm. 9 except (viii) hold, that* $\hat{f}_0 \perp\!\!\!\perp Y_{1:n} \mid X_{1:n}, T_{1:n}$, *and that* $(\mathbb{E}[(\hat{f}_0(X) - \tilde{f}_0(X))^2])^{1/2} = O(1/\sqrt{n})$ *for some fixed* $\tilde{f}_0$. *Then the following three results hold:*

*(a)* *If* $\tilde{f}_0 = f_0$:                    $\hat{\tau}_{W,\hat{f}_0} - \mathrm{SATT} = o_p(1)$.
*(b)* *If* $\|[\tilde{f}_0]\|, \|[f_0]\| < \infty$:       $\hat{\tau}_{W,\hat{f}_0} - \mathrm{SATT} = o_p(1)$.
*(c)* *If* $\|[f_0]\| < \infty, \|[\hat{f}_0]\| = O_p(1)$:   $\hat{\tau}_{W,\hat{f}_0} - \mathrm{SATT} = o_p(1)$.

Note that the above requires $\hat{f}_0$ to be fit on a separate sample to ensure independence. In practice, we simply fit $\hat{f}_0$ in-sample as in Exs. 1 and 2 and in our investigation of performance on real data in Sec. 4.7.

The consistency results for both $\hat{\tau}_W$ and $\hat{\tau}_{W,\hat{f}_0}$ are stronger in the case of KOM, which we discuss next.

## 4. Kernel Optimal Matching.
In this section we develop kernel optimal matching (KOM) methods, which are given by GOM using a reproducing kernel Hilbert space (RKHS). Kernels are standard in machine learning as ways to generalize the structure of learned conditional expectation functions, like classifiers or regressors [30]. Kernels have many applications in statistics, applied and theoretical [31, 32, 33, 34, 35].

A positive semidefinite (PSD) kernel on $\mathcal{X}$ is a function $\mathcal{K} : \mathcal{X} \times \mathcal{X} \to \mathbb{R}$ such that for any $m, x_1, \ldots, x_m$ the Gram matrix $K_{ij} = \mathcal{K}(x_i, x_j)$ is PSD, *i.e.*, $K \in \mathcal{S}_+^{m \times m} = \{A \in \mathbb{R}^{m \times m} : A = A^T, v^T A v \ge 0 \; \forall v \in \mathbb{R}^m\}$. An RKHS on $\mathcal{X}$ is a Hilbert space of functions $\mathcal{X} \to \mathbb{R}$ for which the maps $\mathcal{F} \to \mathbb{R} : f \mapsto f(x)$ are continuous for any $x$. Continuity and the Riesz representation theorem imply that for each $x \in \mathcal{X}$ there is $\mathcal{K}(x, \cdot) \in \mathcal{F}$ such that $\langle \mathcal{K}(x, \cdot), f(\cdot) \rangle = f(x)$ for every $f \in \mathcal{F}$. This $\mathcal{K}$ is always a PSD kernel and reproduces $\mathcal{F}$ in that $\mathcal{F} = \text{closure}\,(\text{span}\,\{\mathcal{K}(x, \cdot) \,:\, x \in \mathcal{X}\})$. Symmetrically, by the Moore-Aronszajn theorem the span of any PSD kernel endowed with $\langle K(x, \cdot), K(x', \cdot)\rangle = K(x, x')$ can be uniquely completed into an RKHS, which we call the RKHS induced by $\mathcal{K}$. PSD kernels also describe the covariance of Gaussian processes: we say that $f \sim \mathcal{GP}(\mu, \mathcal{K})$ if for any $m, x_1, \ldots, x_m$, $(f(x_1), \ldots, f(x_m))$ are jointly normal and $\mathbb{E} f(x_i) = \mu(x_i)$, $\text{Cov}(f(x_i), f(x_j)) = \mathcal{K}(x_i, x_j)$.

Fig 2: Random functions drawn from a Gaussian process

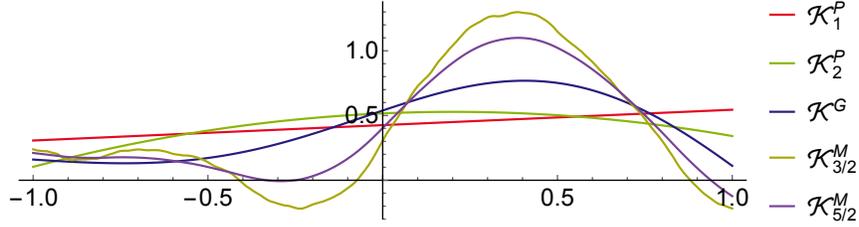

Popular examples of kernels on $\mathbb{R}^d$ are polynomial $\mathcal{K}^{\text{P}}_\nu(x, x') = (1 + \frac{x^T x'}{\nu})^\nu$, exponential $\mathcal{K}^{\text{E}}(x, x') = e^{x^T x'}$, Gaussian $\mathcal{K}^{\text{G}}(x, x') = e^{-\frac{1}{2}\|x - x'\|^2}$, and Matérn $\mathcal{K}^{\text{M}}_\nu(x, x') = \frac{(\sqrt{2\nu}\|x - x'\|)^\nu}{2^{\nu-1}\Gamma(\nu)}\,\text{BK}_\nu(\sqrt{2\nu}\|x - x'\|)$ where $\text{BK}_\nu$ is a modified Bessel function of the second kind. For $s = \nu + d/2 \in \mathbb{N}_0$, the Matérn kernel induces a norm that is equivalent to the Sobolev norm of order $s$ given by $\|f\|^2_{H^s} = \sum_{\alpha \in \mathbb{N}^d_0 : \|\alpha\|_1 \le s} \int_{\mathbb{R}^d} (D^\alpha f)^2$ [36, Cor. 10.13]. More generally, any differential norm $\|f\|^2 = \sum_{\alpha \in \mathbb{N}^d_0} a_{\|\alpha\|_1} \int_{\mathbb{R}^d} (D^\alpha f)^2$ with $a_0 > 0, a_{\lceil (d+1)/2 \rceil} > 0$ also corresponds to an RKHS norm [14, Sec. 6.2.1]. We treat the case of purely differential regularization ($a_0 = 0$) in Sec. 4.5. Fig. 2 displays random draws (on the same realization path) of functions $\mathbb{R} \to \mathbb{R}$ from the Gaussian processes with mean zero and the kernels above.

Generally, we either normalize covariate data before putting it in a kernel so that the control sample has zero sample mean and identity covariance (and choose a length-scale to rescale the norms in the Gaussian and Matérn kernels) or we just fit a rescaling matrix to the data. Much more discussion on this process is given in Sec. 4.6 and on connections to equal percent bias reduction (EPBR) in Appendix A.

RKHS norms always satisfy the conditions of Def. 1 by definition. We call the resulting GOM, kernel optimal matching (KOM).

**Definition 4.** Given a PSD kernel $\mathcal{K}$ on $\mathcal{X}$, the kernel optimal matching $\text{KOM}(\mathcal{W}, \mathcal{K}, \lambda)$ is given by $\text{GOM}(\mathcal{W}, \|\cdot\|, \lambda)$ where $\|\cdot\|$ is the RKHS norm induced by $\mathcal{K}$. We also overload our notation: $\mathfrak{B}(W;\mathcal{K}) = \mathfrak{B}(W;\|\cdot\|)$, $\mathfrak{E}(W;\mathcal{K},\lambda) = \mathfrak{E}(W;\|\cdot\|,\lambda)$.

THEOREM 11. *Let $\mathcal{K}$ be a PSD kernel and let $K_{ij} = \mathcal{K}(X_i,X_j)$. Then, $\mathrm{KOM}(\mathcal{W},\mathcal{K},\lambda)$ is given by the optimization problem*

$$(4.1) \qquad \min_{W \in \mathcal{W}} \ \frac{1}{n_1^2}e_{n_1}^T K_{\mathcal{T}_1,\mathcal{T}_1}e_{n_1} - \frac{2}{n_1}e_{n_1}^T K_{\mathcal{T}_1 \mathcal{T}_0}W + W^T(K_{\mathcal{T}_0 \mathcal{T}_0} + \lambda I)W.$$

Problem (4.1) has a convex-quadratic objective. When $\mathcal{W} = \mathcal{W}^{\mathrm{simplex}}$, the problem is a linearly-constrained quadratic optimization problem, which is polynomially time solvable and easily computed with off-the-shelf solvers.

For subset-based matching $\mathrm{KOM}(\mathcal{W}^{\mathrm{subsets}},\mathcal{K},\lambda)$ is given by the following convex-quadratic binary optimization problem for some $n(\lambda)$:

$$\min_{U \in \{0,1\}^{\mathcal{T}_0}: \sum_{i \in \mathcal{T}_0}U_i = n(\lambda)} \ \frac{1}{n_1^2}e_{n_1}^T K_{\mathcal{T}_1,\mathcal{T}_1}e_{n_1} - \frac{2}{n_1 n(\lambda)}e_{n_1}^T K_{\mathcal{T}_1 \mathcal{T}_0}U + \frac{1}{n(\lambda)^2}U^T K_{\mathcal{T}_0 \mathcal{T}_0}U.$$

Unless otherwise noted, when referring to KOM, we mean on the simplex.

In Cor. 4, we saw that GOM immediately leads to a bound on CMSE. For the case of KOM, we can also interpret it as Bayesian efficient, exactly minimizing the posterior CMSE of $\hat{\tau}_W$, rather than merely bounding it.

THEOREM 12. *Let $\mathcal{K}$ be a PSD kernel, $c \in \mathbb{R}$, $\gamma^2,\sigma^2 \geq 0$, $\lambda = \sigma^2/\gamma^2$. Suppose our prior is that $f_0 \sim \mathcal{GP}(c,\gamma^2\mathcal{K})$ and $Y_i(0) \sim \mathcal{N}(f_0(X_i),\sigma^2)$. Then*

$$\mathbb{E}[(\hat{\tau}_W - \mathrm{SATT})^2 \mid X_{1:n},T_{1:n}] = \gamma^2(\mathfrak{E}(W;\mathcal{K},\lambda) + \lambda/n_1).$$

*If instead our prior is $f_0 \sim \mathcal{GP}(\hat{f}_0,\gamma^2\mathcal{K})$[10] then*

$$\mathbb{E}[(\hat{\tau}_{W,\hat{f}_0}^2 - \mathrm{SATT})^2 \mid X_{1:n},T_{1:n}] = \gamma^2(\mathfrak{E}(W;\mathcal{K},\lambda) + \lambda/n_1).$$

4.1. *Automatic Selection of $\lambda$.* An important question that remains is how to choose $\lambda$. Using the Bayesian interpretation of KOM given by Thm. 12, we treat the choice of $\lambda$ as hyperparameter estimation problem for the prior and employ an empirical Bayes approach.

Given a kernel $\mathcal{K}$, we postulate $f_0 \sim \mathcal{GP}(c,\gamma^2\mathcal{K})$, $Y_i \sim \mathcal{N}(f_0(X_i),\sigma^2)$, where we set $c = \overline{Y}_{\mathcal{T}_0}$. Given this model, we can ask what is the likelihood of the data for a given assignment to $\gamma^2,\sigma^2$. This is known as the *marginal likelihood* as it marginalizes over the actual regression function $f_0$ rather than asking what is the likelihood under a particular choice thereof (as in

---

[10]For example, the posterior of a Gaussian process after observing one fold of the data.

Fig 3: The Balance-Variance Trade-off in KOM

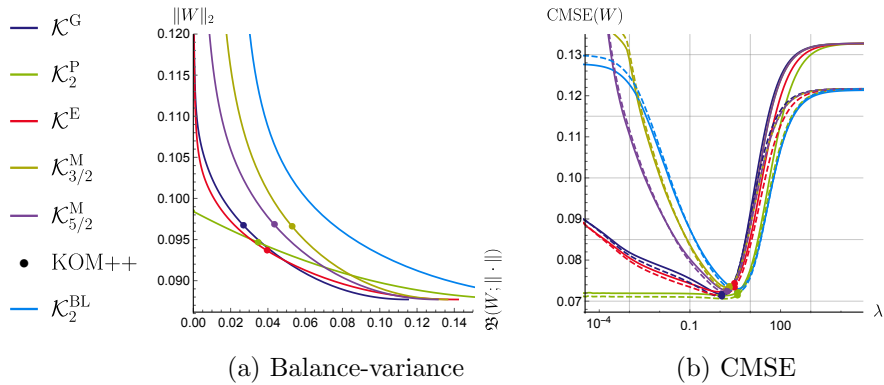

(a) Balance-variance

(b) CMSE

MLE). It is straightforward to show that the negative log marginal likelihood of the control data given the prior parameters $\gamma^2$, $\sigma^2$ is

$$\ell(\gamma^2, \sigma^2) = -\log \mathbb{P}\left(Y_{\mathcal{T}_0} \mid X_{\mathcal{T}_0}, \gamma^2, \sigma^2\right)$$
$$= \tfrac{1}{2}(Y_{\mathcal{T}_0} - \overline{Y}_{\mathcal{T}_0})^T(\gamma^2 K + \sigma^2 I)^{-1}(Y_{\mathcal{T}_0} - \overline{Y}_{\mathcal{T}_0}) + \tfrac{1}{2}\log|\gamma^2 K + \sigma^2 I| + \tfrac{n_0 \log(2\pi)}{2},$$

where $K$ is the Gram matrix on $\mathcal{T}_0$. Choosing $\hat{\gamma}^2, \hat{\sigma}^2$ to minimize this quantity, we let $\hat{\lambda} = \hat{\sigma}^2/\hat{\gamma}^2$. We give the name KOM++ to KOM with $\lambda = \hat{\lambda}$.

In a strict sense, KOM++ does not produce "honest" weights because $\hat{\lambda}$ depends on $Y_{\mathcal{T}_0}$. However, as we only extract a single parameter $\hat{\lambda}$, we guard against data mining, as seen by the resulting low CMSE in the next example and in our investigation of its performance on real data in Sec. 4.7.

EXAMPLE 2.  Let us revisit Ex. 1 to study KOM. We consider KOM with the Gaussian, quadratic, exponential, Matérn $\nu = 3/2$, and Matérn $\nu = 5/2$ kernels. We plot the resulting balance-variance landscape in Fig. 3a. Note that the horizontal axis is different across the curves and so the curves are not immediately comparable. We plot the resulting CMSE in Fig. 3b. In both figures, we point out the result of KOM++, which chooses $\lambda$ by marginal likelihood and which appears to perform well across all kernels. We include SKOM with the second-order Beppo-Levi kernel $\mathcal{K}_2^{\mathrm{BL}}$ as detailed in Sec. 4.5.

4.2. *Consistency.* We now characterize conditions for KOM to lead to consistent estimation. Under correct specification, we can guarantee efficient $\sqrt{n}$-consistency. Under incorrect specification but with a $C_0$-universal kernel, we can still ensure consistency. A $C_0$-universal kernel, defined below, is one that can arbitrarily approximate compactly-supported continuous functions in $L_\infty$. The Gaussian and Matérn kernels are $C_0$-universal and the exponential kernel is $C_0$-universal on compact spaces [37].

DEFINITION 5. A PSD kernel $\mathcal{K}$ on a Hausdorff $\mathcal{X}$[11] is $C_0$-universal if, for any continuous function $g : \mathcal{X} \to \mathbb{R}$ with compact support (i.e., for some $C$ compact, $\{x : g(x) \neq 0\} \subseteq C$) and $\eta > 0$, there exists $m, \alpha_1, x_1, \ldots, \alpha_m, x_m$ such that $\sup_{x \in \mathcal{X}} |\sum_{j=1}^m \alpha_i \mathcal{K}(x_j, x) - g(x)| \leq \eta$.

THEOREM 13. *Suppose Asns. 1 and 2 hold and that*

(i)    *for each $n$, $W$ is given by* $\mathrm{KOM}(\mathcal{W}, \mathcal{K}, \lambda_n)$,
(ii)   $\lambda_n \in [\underline{\lambda}, \overline{\lambda}] \subset (0, \infty)$,
(iii)  $\mathcal{W}^{subsets} \subseteq \mathcal{W}$,
(iv)   $\mathbb{E}[\mathcal{K}(X, X) \mid T = 1] < \infty$, *and*
(v)    $\mathrm{Var}(Y(0) \mid X)$ *is almost surely bounded.*

*Then the following two results hold:*

(a)   *If $\|[f_0]\| < \infty$:*          $\hat{\tau}_W - \mathrm{SATT} = O_p(n^{-1/2})$.
(b)   *If $\mathcal{K}$ is $C_0$-universal:*   $\hat{\tau}_W - \mathrm{SATT} = o_p(1)$.

As before, condition (iii) is satisfied for subset, mutlisubset, and simplex matching. Condition (iv) is trivially satisfied for any bounded kernel ($\mathcal{K}(x, x) \leq M$), such as the Gaussian and Matérn kernels. Condition (v) is generally weak and, in particular, is satisfied under homoskedasticity. Case (a) is the case of a well-specified model, even if $\mathcal{K}$ induces an infinite-dimensional RKHS (all $C_0$-universal kernels do). For example, while the exponential kernel is infinite dimensional and $C_0$-universal on compact spaces, polynomial functions (e.g., linear) have finite norm in its induced RKHS. Moreover, under the common semiparametric specification where $f_0$ is assumed to be Sobolev (e.g., be square-integrable so that $\mathrm{Var}(Y(0)) < \infty$ and have square-integrable derivatives of degrees up to $\lceil (d+1)/2 \rceil$), it is well-specified by the Matérn kernel. Case (b) is the case of a misspecified model, wherein a $C_0$-universal kernel still guarantees model-free consistency.

The results can be extended to the AW estimator, which we term the augmented kernel weighted (AKW) estimator when combined with KOM.

THEOREM 14. *Suppose the conditions of Thm. 13 hold and that $\hat{f}_0 \perp\!\!\!\perp Y_{1:n} \mid X_{1:n}, T_{1:n}$. Then the following five results hold:*

---

[11]Euclidean space $\mathbb{R}^d$, for example, is Hausdorff.

(a) If $\|[\hat{f}_0 - f_0]\| = o_p(1)$:
$$\hat{\tau}_{W,\hat{f}_0} - \text{SATT} = \frac{1}{n_1}\sum_{i \in \mathcal{T}_1}\epsilon_i - \sum_{i \in \mathcal{T}_0}W_i\epsilon_i + o_p(n^{-1/2}).$$

(b) If $\|[f_0]\| < \infty, \|[\hat{f}_0]\| = O_p(1)$:   $\hat{\tau}_{W,\hat{f}_0} - \text{SATT} = O_p(n^{-1/2}).$

If $(\mathbb{E}[(\hat{f}_0(X) - \tilde{f}_0(X))^2])^{1/2} = O(r(n))$ for $r(n) = o(1)$ and

(c) If $\tilde{f}_0 = f_0$:   $\hat{\tau}_{W,\hat{f}_0} - \text{SATT} = O_p(r(n) + n^{-1/2}).$

(d) If $\mathcal{K}$ is $C_0$-universal:   $\hat{\tau}_{W,\hat{f}_0} - \text{SATT} = o_p(1).$

(e) If $\|[\tilde{f}_0]\|, \|[f_0]\| < \infty$:   $\hat{\tau}_{W,\hat{f}_0} - \text{SATT} = O_p(r(n) + n^{-1/2}).$

Case (c) says that $\hat{f}_0$ is weakly $L_2$-consistent with rate $r(n)$. The condition holds with a parametric $r(n) = n^{-1/2}$, for example, if $f_0$ is linear and $\hat{f}_0$ is given by OLS on $X_{\mathcal{T}_0}, Y_{\mathcal{T}_0}$. Cases (d), (e), and (b) say that when $\hat{f}_0$ may be inconsistent, kernel weights can correct for the error, with the situation varying on whether the regression estimator is itself in the RKHS. Thus, one useful case is when $\hat{f}_0$ is given by a parametric regression and we use KOM with a $C_0$-universal kernel: then we get a parametric rate when the model is correct but do not sacrifice consistency when it is not. Case (a) shows that if $\hat{f}_0$ is consistent in the RKHS norm then we are only left with the efficient irreducible error that involves only the residuals $\epsilon_i$, which of course $X$ cannot control for. Example of such cases include when $\hat{f}_0$ is given by a well-specified kernel ridge regression and we use KOM with the same kernel or when its given by any nonparametric regression and its derivatives up to $\lceil (d+1)/2 \rceil$ are weakly consistent and we use KOM with the Matérn kernel.

For the case where $\hat{f}_0$ is given by kernel ridge regression and $W$ by KOM with $\mathcal{W}^{\text{general}}$ (see Sec. 5.6) and both use the same kernel and $\lambda$, we get a closed form for AKW:

$$\frac{1}{n_1}e_{n_1}^T(Y_{\mathcal{T}_1} - K_{\mathcal{T}_1\mathcal{T}_0}(K_{\mathcal{T}_0\mathcal{T}_0} + \lambda I)^{-1}(K_{\mathcal{T}_0\mathcal{T}_0} + 2\lambda I)(K_{\mathcal{T}_0\mathcal{T}_0} + \lambda I)^{-1}Y_{\mathcal{T}_0}).$$

This estimator essentially debiases the kernel ridge regression adjustment – bias which is unavoidable in nonparametric kernel ridge regression (*e.g.*, universal kernel). In the parametric case (rank of $K_{\mathcal{T}_0\mathcal{T}_0}$ is bounded), we can set $\lambda = 0$ and recover plain OLS adjustment as it is already unbiased.

In a related alternative usage of AW estimators, [38] recently showed that in a setting with a *correctly specified* but *high-dimensional* parametric (linear) model, efficient estimation is possible using $\hat{\tau}_{W,\hat{f}_0}$ with $\hat{f}_0$ given by LASSO [39] and $W$ given by the equivalent of GOM($\mathcal{W}^{\text{simplex}}, \|\cdot\|_{1\text{-lin}}, \lambda$).

4.3. *Kernel Matching to Reduce Model Dependence.* A popular use of matching, as implemented in the popular $R$ package `MatchIt`, is as preprocessing before regression analysis, in which case matching is commonly understood to reduce model dependence [4]. Similar in spirit to double robustness in the face of potential model misspecification, this is understood commonly and in [4] as pruning unmatched control subjects before a linear regression-based treatment effect estimation. More generally, however, we can consider any nonnegative weights $W$, whether subset or simplex weights. This leads to the following weighted least squares estimator:

$$\hat{\tau}_{\mathrm{WLS}(W)} = \operatorname{argmin}_{\tau \in \mathbb{R}} \min_{\alpha \in \mathbb{R}, \beta_1, \beta_2 \in \mathbb{R}^d} \sum_{i=1}^{n} (T_i/n_1 + (1 - T_i)W_i)$$
$$\times (Y_i - \alpha - \tau T_i - \beta_1^T X_i - \beta_2^T (X_i - \overline{X}_{\mathcal{T}_1})T_i)^2$$

When $W$ is given by KOM, we can show that this procedure indeed achieves the desired robustness: consistency without model dependence and parametric rates when the model is correctly specified.

THEOREM 15.   *Suppose the conditions of Thm. 13 hold and that $\mathcal{K}$ is $C_0$-universal, $\mathcal{X}$ bounded, and $\mathbb{E}[XX^T \mid T = 1]$ non-singular. Then,*

(a)  *Regardless of $f_0$:*                        $\hat{\tau}_{\mathrm{WLS}(W)} - \mathrm{SATT} = o_p(1).$
(b)  *If $\exists \alpha_0, \beta_0$ s.t. $f_0(x) = \alpha_0 + \beta_0^T x$:*   $\hat{\tau}_{\mathrm{WLS}(W)} - \mathrm{SATT} = O_p(n^{-1/2}).$

4.4. *Inference and Partial Identification.*   In order to conduct inference on the value of SATT using KOM, it is important to develop appropriate standard errors or other confidence intervals for $\hat{\tau}_W$. There are several options for estimating standard errors. One general-purpose option is the bootstrap. In applying the bootstrap to KOM, we re-optimize the weights for each bootstrap sample and record the resulting estimator to produce the bootstrap distribution (rather than, say, using a weighting function precomputed at the onset on the complete dataset). Quantile, studentized, and $\mathrm{BC_A}$ bootstrap intervals are possible choices [40]. Another general-purpose option is to use the estimate $\hat{\tau}_{\mathrm{WLS}(W)}$ and employ the corresponding robust sandwich (Huber-White) standard errors.

However, more specialized procedures are possible. One particularly appealing nature of matching is the transparent structure of the data [41, Ch. 6]: it preserves the unit of analysis since the result, like the raw control sample, is still a valid distribution over the control units, whether it is a subset, a multisubset with duplicates, or any redistribution that is nonnegative and sums to one. This is preserved in KOM and enables the use of similar inferential methods as used in standard matching that interpret the data as a weighted sample.[12] In particular, since the weighted estimator $\hat{\tau}_W$

---

[12]KOM, however, does not preserve the property of producing finitely-many coarsened strata of units as in such methods as full matching.

Fig 4: Inference with KOM++

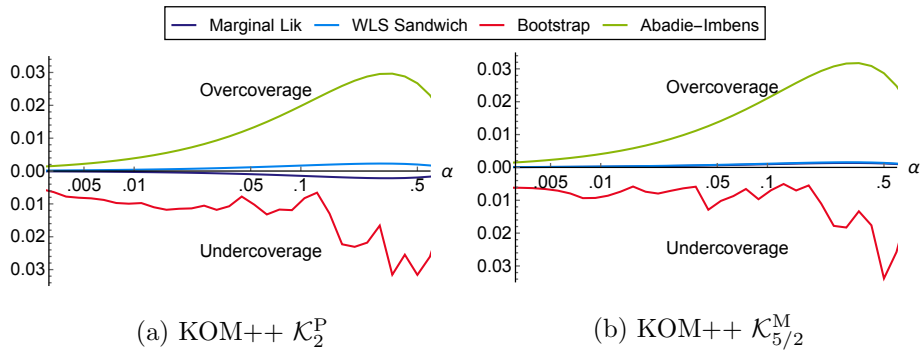

(a) KOM++ $\mathcal{K}_2^{\mathrm{P}}$                    (b) KOM++ $\mathcal{K}_{5/2}^{\mathrm{M}}$

for SATT exactly matches the criteria presented in [15, §19.8], one approach to compute standard errors is to use the within-treatment-group matching techniques developed in [42, 43, 11] to estimate residual variances. In the specific case of KOM++, one can also use the marginal likelihood estimate of the residual variance to produce a standard error based on Thm. 1.[13] The next example explores how to use these methods to produce confidence intervals for KOM. Then, we will see how the interpretable nature of KOM can also allow us to account for unavoidable imbalances that lead to deceptive point estimates and instead produce more honest interval estimates.

EXAMPLE 3. We revisit Ex. 2 to look at (finite-sample) inference on SATT using KOM++ based on each of the confidence intervals above. We consider the situation of a constant effect $Y(1) - Y(0) = \tau$ and plot the desired significance $\alpha$ against the difference of the actual coverage and $1 - \alpha$ in Fig. 4 for two examples: quadratic and Matérn. Coverage is computed keeping $X_{1:n}, T_{1:n}$ fixed. A conservative confidence interval corresponds to a point above the horizontal axis. For the method of [15, §19.6], we use a single match. Given an estimate $\hat{\tau}$, standard error estimate $\hat{s}$, and desired significance $\alpha$, we construct the confidence region $\hat{\tau} \pm \Phi^{-1}(1 - \alpha/2)\hat{s}$. For the bootstrap, we construct the confidence interval as the interval between the $\alpha/2$ and $1 - \alpha/2$ quantiles of the bootstrap distribution over 1000 re-samples.

All above confidence interval methods discard the conditional bias term in the error of $\hat{\tau}_W$. To quote [15], "with a sufficiently flexible estimator, this term will generally be small," meaning asymptotically insignificant compared to standard error. Indeed, in the above example, bias is small and estimated standard errors alone achieve approximately valid confidence intervals. However, in settings where overlap is limited, the bias may be in fact be significant. In the extreme case of no overlap (Asn. 2 does not hold), causal effects may be unidentifiable and bias may be unavoidable. In settings of low overlap, standard inverse propensity weighting approaches lead to very large weights and high variance and, in the extreme case of no overlap, they lead to infinite weights and provide no insights into average causal effects (unless we change the target of estimation as in [45]). Similarly, matching methods will fail to find good matches and a significant bias will remain.

---

[13]Marginal likelihood can also be used to estimate heteroskedastic noise [44].

KOM opens the door to the possibility of partial identification of causal effects in the absence of overlap by bounding or approximately bounding the bias. With KOM, we can obtain an explicit bound on the bias: it is bounded by $\|[f_0]\|\mathfrak{B}(W;\mathcal{K})$. We may have an a priori bound on $\|[f_0]\| \leq \hat{\gamma}$. For example, using the Beppo-Levi kernel presented in the next section, this can take the form of an a priori bound on the roughness of $f_0$. Alternatively, in the case of KOM++, we may take a data-driven approach and rely on the marginal likelihood estimate $\hat{\gamma}$ of $\|[f_0]\|$. That is, in the absence of overlap, we assume we can still judge the complexity of $f_0$ on the treated population of $X$ – instead of the actual values – by observing its values with noise on the control population of $X$. In either case, assuming that $\|[f_0]\| < \infty$, we can obtain an *interval estimate* of the treatment effect:

$$(4.2) \qquad \hat{\tilde{\mathfrak{T}}}_W = [\hat{\tau}_W - \hat{\gamma}\mathfrak{B}(W;\mathcal{K}), \ \hat{\tau}_W + \hat{\gamma}\mathfrak{B}(W;\mathcal{K})].$$

This interval accounts for the possible bias precisely in terms of the covariate imbalances left in the reweighted samples and in the extent to which $f_0$ could, in the worst case, depend on these imbalances and induce the worst bias. If this characterization of $f_0$ is valid (*i.e.*, $\|[f_0]\| \leq \hat{\gamma}$) then this interval contains SATT, as a trivial consequence of Thm. 1. To the interval estimate in eq. (4.2), we can further add standard errors to produce a robust confidence interval that provides coverage even in cases of limited overlap.

That KOM preserves the unit of analysis is critical. Using negative weights that do not necessarily sum to one, one can always make perfectly zero any imbalance metric, including that used by KOM – the result is essentially equivalent to regression (see Sec. 5.6). This, however, requires extreme extrapolation and the sense in which this is a "perfect match" is highly deceptive: in the absence of overlap and parametric specification, identification is simply impossible. The result of using such weights would be largely meaningless. Instead, KOM avoids extrapolation by preserving the unit of analysis and restricts only to a valid distribution over the controls (potentially, if so restricted, to a proper subset of the control sample). The imbalances between

Fig 5: Partial Identification with KOM++

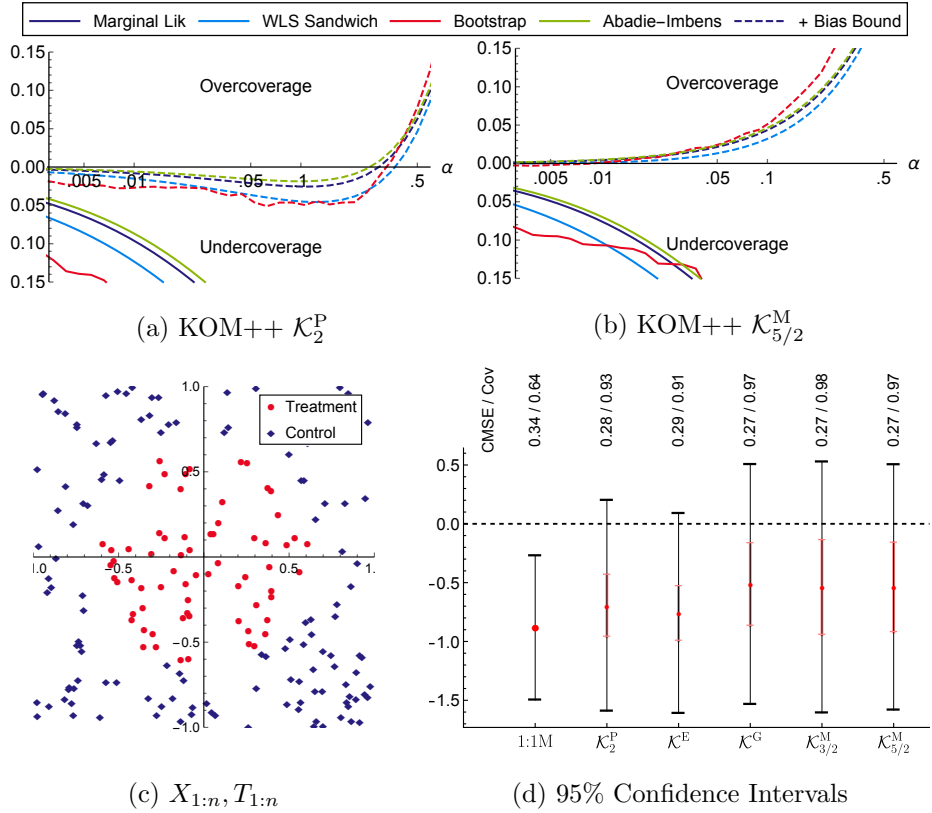

(a) KOM++ $\mathcal{K}_2^{\mathrm{P}}$

(b) KOM++ $\mathcal{K}_{5/2}^{\mathrm{M}}$

(c) $X_{1:n}, T_{1:n}$

(d) 95% Confidence Intervals

this distribution of controls and the empirical distribution of treated units are transparent and easily red off from the result of the KOM optimization problem. This enables us to construct an honest and robust interval for the effect, without extrapolating beyond what we actually observe.

EXAMPLE 4. We repeat Ex. 3 but change the distribution of covariates to eliminate overlap completely. Instead of drawing $T$ as given by $\mathbb{P}(T = 1 \mid X)$ in Ex. 1, we fix $T_i = 1$ whenever this given propensity function is greater than 0.4 and otherwise $T_i = 0$. This yields the draw in Fig. 5c, which has $n_0 = 133, n_1 = 67$. We repeat Ex. 3 either as before (solid lines) or by instead adding the confidence terms to the endpoints of the interval estimate in eq. (4.2) using the $\hat{\gamma}$ estimate from the application of KOM++ (dashed lines) and plot the coverage in Figs. 5a and 5b. In Fig. 5d, we compare the point estimate given by 1:1M (red dot) for SATT (dashed line) to the interval estimate given by KOM++ (pink intervals), both surrounded by a confidence interval given by the standard error of [15, §19.6]. The CMSE (above plot) of the KOM++ point estimates is not a substantial improvement over 1:1M – little can be done in the face of such an extreme lack of overlap – but the robust confidence intervals that arise from accounting for the bias help in achieving correct coverage (above plot) in the face of lack of overlap.

4.5. *Semi-Kernel Optimal Matching.* We next extend KOM to the semiparametric case with unconstrained parametric part, where we combine both a parametric exact matching criterion such as exact matching of means with a non-parametric criterion such as that of KOM. A notable example will include matching against all functions with square-integrable Hessians, as in smoothing splines [46, Sec. 5.7]. First, we define *conditionally* PSD kernels.

For a class of functions $\mathcal{G} \subseteq [\mathcal{X} \to \mathbb{R}]$, a $\mathcal{G}$-conditionally PSD kernel on $\mathcal{X} \subset \mathbb{R}^d$ is a symmetric function $\mathcal{K} : \mathcal{X} \times \mathcal{X} \to \mathbb{R}$ that satisfies $\sum_{i=1}^{m} v_i v_j \mathcal{K}(x_i, x_j) \geq 0$ for every $m, x_1, \ldots, x_m, v_1, \ldots, v_m$ satisfying $\sum_{i=1}^{n} v_i g(x_i) = 0$ for all $g \in \mathcal{G}$. For example, $\{0\}$-conditionally PSD kernels are just the PSD kernels. Given a $\mathcal{G}$-conditionally PSD kernel $\mathcal{K}$, we can define a corresponding magnitude:

$$(4.3) \quad \|f\|^2 = \inf \left\{ \sum_{i,j=1}^{\infty} \alpha_i \alpha_j \mathcal{K}(x_i, x_j) : \begin{array}{l} f = g + \sum_{i=1}^{\infty} \alpha_i \mathcal{K}(x_i, \cdot), \\ g \in \mathcal{G}, \sum_{i=1}^{\infty} \alpha_i^2 \mathcal{K}(x_i, x_i) < \infty, \\ \sum_{i=1}^{\infty} \alpha_i g'(x_i) = 0 \ \forall g' \in \mathcal{G} \end{array} \right\}.$$

If $\mathcal{K}$ is $\mathcal{G}$-conditionally PSD, then we refer to GOM($\mathcal{W}, \|\cdot\|, \lambda$) with $\|\cdot\|$ as in eq. (4.3) as SKOM($\mathcal{W}, \mathcal{K}, \lambda$), abbreviating SKOM for semi-kernel optimal matching and treating $\mathcal{G}$ as implicit in $\mathcal{K}$.

An important example is smooth functions on $\mathbb{R}^d$. Let $\mathcal{G}$ be all polynomials of degree at most $\nu - 1$: $\mathcal{G}_\nu^{\text{poly}} = \text{span}\{x^\alpha : \alpha \in \mathbb{N}_0^d, \|\alpha\|_1 \leq \nu - 1\}$. Then, For $\nu > d/2$, the Beppo-Levi kernel $\mathcal{K}_\nu^{\text{BL}}(x, x') = \kappa_\nu(\|x - x'\|_2)$ is $\mathcal{G}_\nu^{\text{poly}}$-conditionally PSD, where $\kappa_\nu(0) = 0$ and $\kappa_\nu(u) = (-1)^{\nu+(d-2)/2} u^{2\nu-d} \log(u)$ for $d$ even and $\kappa_{d,\nu}(u) = u^{2\nu-d}$ for $d$ odd. The Beppo-Levi kernel's corresponding magnitude in eq. (4.3) is equivalent to the square-integral of the $\nu^{\text{th}}$ derivatives [36, Prop. 10.39]: $\|f\|_{\text{BL}}^2 = \sum_{\alpha \in \mathbb{N}_0^d, \|\alpha\|_1 = \nu} \binom{\nu}{\alpha} \int_{\mathbb{R}^d} (D^\alpha f)^2$.

Two important cases are cubic and thin-plate splines. In $d = 1$, the cubic spline kernel $\mathcal{K}_2^{\text{BL}}(x, x') = |x - x'|^3$ is conditionally PSD with respect to all linear functions $\mathcal{G}^{\text{lin}} = \mathcal{G}_2^{\text{poly}}$. In $d = 2$, the thin-plate spline kernel $\mathcal{K}_2^{\text{BL}}(x, x') = \|x - x'\|^2 \log(\|x - x'\|)$ is also $\mathcal{G}^{\text{lin}}$-conditionally PSD. In either case, the corresponding magnitude in eq. (4.3) is equivalent to the roughness of $f$, or square integral of the Hessian:

$$\|f\|_{\text{Roughness}}^2 = \int_{\mathbb{R}^d} \|\nabla^2 f\|_{\text{Frobenius}}^2.$$

Thus, SKOM$(\mathcal{W}, \mathcal{K}_2^{\mathrm{BL}}, \lambda)$ seeks weights to balance *all* smooth functions. In particular, as the linear part of $f$ is completely unconstrained since linear functions have zero Hessian, it will, if possible, *exactly* balance all linear functions, *i.e.*, it will exactly match the sample means.

Like KOM, SKOM admits a solution as a convex-quadratic-objective optimization problem.

THEOREM 16. *Let $\mathcal{G} = \mathrm{span}\{g_1, \ldots, g_m\}$, let $\mathcal{K}$ be a $\mathcal{G}$-conditionally PSD kernel, let $K_{ij} = \mathcal{K}(X_i, X_j)$, let $G_{ij} = g_i(X_j)$, and let $N \in \mathbb{R}^{n \times k}$ have columns forming a basis for the null space of $G$. Then SKOM$(\mathcal{W}, \mathcal{K}, \lambda)$ is given by $W = N_{\mathcal{T}_0} U$ with $U$ given by the optimization problem*

$$(4.4) \qquad \begin{aligned} \min \quad & U^T N^T (K + \lambda I_{\mathcal{T}_0}) N U \\ s.t. \quad & U \in \mathbb{R}^k, \; N_{\mathcal{T}_0} U \in \mathcal{W}, \; N_{\mathcal{T}_1} U = -e_{n_1}/n_1 \end{aligned}$$

*where $(I_{\mathcal{T}_0})_{ij} = \mathbb{I}[i = j \in \mathcal{T}_0]$. Moreover, $N^T K N$ is a PSD matrix so that the quadratic objective is necessarily convex.*

Note that in the case of $\mathcal{W} = \mathcal{W}^{\mathrm{simplex}}$, if a constant function is in $\mathcal{G}$ as in the case of smooth functions, then the constraint $\sum_{i=1}^n W_i = 1$ is redundant in problem (4.4) as it is already enforced by the other constraints. In particular, the constraints necessarily imply $B(W; g) = 0$ for all $g \in \mathcal{G}$.

This also means that, in the case of SKOM over smooth functions with $\mathcal{W} = \mathcal{W}^{\mathrm{simplex}}$, there may exist *no* solution at all to the SKOM unless the treated sample mean $\overline{X}_{\mathcal{T}_1} = \frac{1}{n_1} \sum_{i \in \mathcal{T}_1} X_i$ is in the convex hull of the control sample conv$\{X_i : i \in \mathcal{T}_0\}$. If it is, then SKOM will seek the weights that simultaneously match the means exactly without extrapolation and balance all smooth functions. If it is not, a solution will nonetheless exist if we instead use $\mathcal{W} = \mathcal{W}^{\mathrm{general}}$ (effectively fit a spline), but this allows extrapolation and is inadvisable. More appropriately, in the case where exactly matching means without extrapolation is not feasible, one should instead seek to achieve approximate matching without extrapolation by simply using standard KOM (which penalizes linear terms in $f_0$). First-order discrepancies can be emphasized by putting higher weight on linear functions, *e.g.*, using a direct sum of a universal RKHS with an appropriately weighted linear RKHS.

4.6. *Automatic Selection of $\mathcal{K}$.* We can go further than just selecting $\lambda$ in a data-driven manner for KOM and also use marginal likelihood to choose $\mathcal{K}$. Consider a parametrized family of kernels $\mathfrak{K} = \{\mathcal{K}_\theta(x, x') : \theta \in \Theta\}$. The most common example is parameterizing the length-scale of the Gaussian kernel: $\{\mathcal{K}_\theta(x, x') = \mathcal{K}^{\mathrm{G}}(x/\theta, y/\theta) : \theta > 0\}$. But we can easily conceive of more complex structures such as fitting a rescaling matrix for any kernel: $\{\mathcal{K}_\theta(x,x') = \mathcal{K}(\theta x, \theta y) : \theta \in \Theta \subseteq \mathbb{R}^{d \times d}\}$, where $\Theta$ can be restricted to diagonal matrices in order to rescale each covariate (known as automatic relevance detection, ARD), can be unrestricted to fit a full covariance structure, or can be restricted to matrices with only $d' < d$ rows in order to find a projection onto a lower dimensional space, which can help make the subsequent matching more efficient. Additionally, we can consider mixtures of kernels, $\{\theta\mathcal{K}_1 + (1-\theta)\mathcal{K}_2 : \theta \in [0,1], \mathcal{K}_1 \in \mathfrak{K}_1, \mathcal{K}_2 \in \mathfrak{K}_2\}$, and more complex structures like the spectral mixture kernel [47].

Table 2

| | $\hat{\tau}_W$ | $\hat{\tau}_{W,\hat{f}_0}$ | $\hat{\tau}_{\text{WLS}(W)}$ |
|---|---|---|---|
| KOM++ ARD $\mathcal{K}_4^P$ | 0.028481 | 0.028698 | 0.028685 |
| KOM++ ARD $\mathcal{K}_{3/2}^M$ | 0.028886 | 0.029165 | 0.029182 |
| KOM++ ARD $\mathcal{K}_{5/2}^M$ | 0.028983 | 0.029279 | 0.029323 |
| KOM++ ARD $\mathcal{K}^G$ | 0.029033 | 0.029288 | 0.029316 |
| KOM++ ARD $\mathcal{K}^E$ | 0.029072 | 0.029188 | 0.029128 |
| KOM++ $\mathcal{K}^G$ | 0.029783 | 0.029834 | 0.029856 |
| KOM++ $\mathcal{K}_{5/2}^M$ | 0.029895 | 0.029935 | 0.029956 |
| KOM++ $\mathcal{K}_{3/2}^M$ | 0.029944 | 0.029980 | 0.030001 |
| KOM++ $\mathcal{K}_4^P$ | 0.030391 | 0.030471 | 0.030543 |
| Inv Prop Weights | 0.033168 | 0.033146 | 0.033126 |
| No matching | 0.034188 | 0.032925 | 0.032925 |
| CEM++ | 0.039811 | 0.039611 | 0.039533 |
| CEM | 0.040418 | 0.040228 | 0.040073 |
| NNM++ | 0.042890 | 0.043184 | 0.043511 |
| NNM | 0.047071 | 0.047399 | 0.047695 |
| PSM | 0.053359 | 0.052221 | 0.052415 |

It is easy to see that given any such parameterized kernel $\mathcal{K}_\theta$, the negative log marginal likelihood is simply given by the parametrized Gram matrix:

$$\ell(\theta, \gamma^2, \sigma^2) = \tfrac{1}{2}(Y_{\mathcal{T}_0} - \overline{Y}_{\mathcal{T}_0})^T (\gamma^2 K_\theta + \sigma^2 I)^{-1} (Y_{\mathcal{T}_0} - \overline{Y}_{\mathcal{T}_0})$$
$$+ \tfrac{1}{2}\log|\gamma^2 K_\theta + \sigma^2 I| + \tfrac{n_0 \log(2\pi)}{2},$$

where $K_{\theta,i,j} = \mathcal{K}_\theta(X_i, X_j)$ for $i, j \in \mathcal{T}_0$. As before, we can optimize this over $\theta, \gamma^2, \sigma^2$ jointly to select both $\mathcal{K}$ and $\lambda$ for KOM.

Note that if it is the case that for any $\theta \in \Theta$ and unitary matrix $U$ we have $\mathcal{K}_\theta(Ux, Ux') = \mathcal{K}_{\theta'}(x, x')$ for some $\theta' \in \Theta$, then KOM after marginal likelihood is affinely invariant. For example, this is the case when we parametrize either an unrestricted low-dimensional projection or full covariance matrix for the kernel. This means that it is not necessary to preprocess the data to make the sample covariance identity by studentization.

EXAMPLE 5. We revise the setup of Ex. 1 with higher dimensions and data size: we let $n = 500$, $X \sim \text{Unif}[-1,1]^5$, $\mathbb{P}(T = 1 \mid X) = 0.95/(1 + \frac{3}{\sqrt{5}}\|X\|_2)$, and $Y(0) \mid X \sim \mathcal{N}(X_1^2 + X_2^2 - X_1/2 - X_2/2, \sqrt{3})$, so that there

Fig 6: Effect Estimation for the Infant Health and Development Program

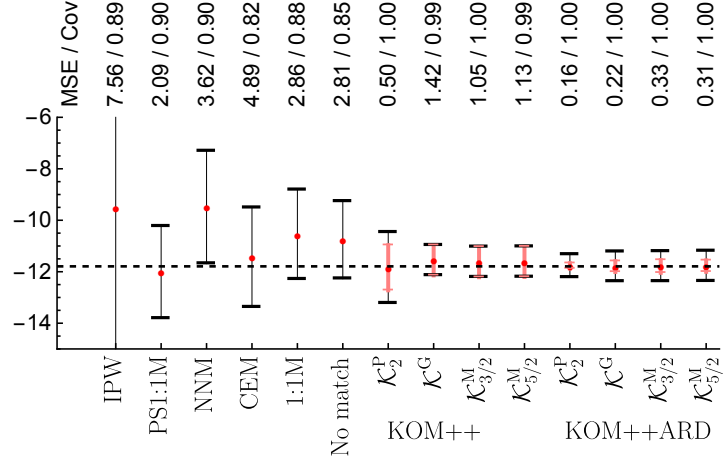

are three redundant covariates. We consider all estimators in the preceding examples alone with KOM++ with ARD and propensity weights (including AIPW) and propensity score (1:1) matching (PSM) with propensities estimated by logistic regression. For CEM, we coarsen into the greatest number of levels per covariate while maintaining at least one control unit in each stratum with a treated unit. For each method, we interpret the result as a set of weights and consider either the simple weighting estimator $\hat{\tau}_W$, the AW estimator $\hat{\tau}_{W,\hat{f}_0}$ with $\hat{f}_0$ given by OLS, and the weighted least squares estimator $\hat{\tau}_{\mathrm{WLS}(W)}$. We run 500 replications and tabulate the *marginal* mean squared error (MSE) for estimating SATT in Tab. 2.

4.7. *Infant Health and Development Program.* We next consider evaluating the practical usefulness of KOM++ by studying data from the Infant Health and Development Program (IHDP). IHDP was a randomized experiment intended to measure the effect of a program consisting of child care and home visits from a trained provider on early child development [48], as measured through cognitive test scores.

To make an observational study from this data, we follow the construction of [49], where the subject of study is a child. We make one modification to further exacerbate overlap. Like [49], we remove all children with non-white mothers from the treatment group. To make overlap worse, we further remove all children with mothers aged 23 or younger from the treatment group and all children with mothers that are either white or aged 26 or older in the control group. In sum, the treatment group ($n_1 = 94$) consists only of children with older white mothers and the control group ($n_0 = 279$) consists only of children with younger nonwhite mothers, creating groups that are highly disparate in socioeconomic privilege. (The age cutoffs are near the mean and were chosen so to keep the data non-linearly-separable or else propensity score methods with scores estimated by logistic regression would be undefined.) Each unit is described by 25 covariates, 6 continuous and 19 binary, corresponding to measurements on the child (birth weight, etc.), measurements on the child's mother (smoked during pregnancy, etc.), and site. We generate outcomes $Y_i(0), Y_i(1)$ precisely as described by the *non-linear* response surface of [49] ("response surface B") with the sole restriction that we condition on the coefficient on mother's age being zero.

We consider a range of methods: standard methods, KOM++ with various kernels, and KOM++ with ARD. The standard methods we consider include: inverse propensity weighting (IPW) and propensity score (1:1) matching (PSM) both using propensity scores estimated by logistic regression, NNM and 1:1M on the Mahalanobis distance, and CEM on a coarsening chosen for maximal overlap. Coarsening on all variables (treating continuous variables as indicators for being above or below the mean) creates too many strata ($2^{25}$) and leaves only one treated unit in a stratum with at least one control unit. Instead, for CEM, we select half of the covariates (13) to maximize the number of treated units that are in a stratum with at least one control unit after coarsening on only these covariates. Following the suggestion of [5], we then proceed to prune all treated units in strata without overlap, leaving 69 units (only for CEM). We also omit NNM++ due to the high computational burden of its cross-validation procedure. For KOM++ we consider the quadratic ($\mathcal{K}_2^{\mathrm{P}}$), Gaussian ($\mathcal{K}^{\mathrm{G}}$), and Matérn ($\mathcal{K}_{3/2}^{\mathrm{M}}, \mathcal{K}_{5/2}^{\mathrm{M}}$) kernels and either using or not using ARD. We compute standard errors for all methods using the method of [15, §19.6]. We construct confidence intervals by adding 1.96 standard errors to either the point estimate for all standard methods or to the interval estimate given by eq. (4.2) for KOM++, using the magnitude of $f_0$ as estimated by the marginal likelihood step.

We plot the results in Fig. 6. At the top, the figure lists the *marginal* mean-squared errors (MSE) $\mathbb{E}[(\hat{\tau} - \mathrm{SATT})^2]$ and the coverage of the 95% confidence intervals over 10,000 runs (note that SATT differs by run). Below the MSEs, the figure shows the results from one representative example run, showing SATT (dashed line), point estimates (red dots), confidence intervals (black bars), and, in the case of KOM++, interval estimates (pink bars). It is clear that among matching and weighting methods, KOM++ and, in particular, KOM++ when using ARD leads to significantly smaller error. As a weighting method, it can easily be combined with any regression technique by reweighting the training data or by reweighting the average of residuals.

The low overlap in this example leads IPW to produce extreme weights and suffer high MSE. In comparison, KOM++, while it cannot fix the unavoidable bias due to lack of overlap in and of itself, is able to maintain stable weights and small MSE by considering error directly while limiting extrapolation. 1:1M also provides rather stable weights but has a much harder time achieving good balance, leading to significantly higher MSE than KOM++. Through the lens of GOM we can identify two causes for this. On the one hand, 1:1M is only a heuristic way to trade off balance and variance: as seen in Ex. 1, it is not necessarily on the efficient frontier. On the other, it is trying to balance far too much than is really necessary. One way to understand 1:1M's slow convergence rate in $d \geq 2$ [11] is that finding good pairs becomes rapidly hard as dimension grows modestly. GOM offers another, functional-analytic perspective: 1:1M, which optimizes balance with respect to Lipschitz functions, is trying to balance far too much. Lipschitz functions are not only infinite dimensional, they are also non-separable, *i.e.*, no countable subset is dense. They have too little structure to be practically useful. In comparison, a $C_0$-universal RKHS, such as those given by the Gaussian or Matérn kernels, can still approximate any function arbitrarily well, but is still a separable space, admitting a countable orthonormal basis so that KOM++ is essentially balancing a countable number of moments. Essentially, this imposes enough structure so that good balance is actually achievable but not too much so that the resulting method remains fully non-parametric. Often, as in this example, even quadratic is enough, but it does not hurt too much to use the Gaussian or Matérn kernels and guarantee consistency without specification. Adding ARD to KOM++ significantly improves its performance by learning the right representation of the data to balance, leading to lower MSEs.

4.8. *Recommendations for Practical Use.* In sum the results, both theoretical and empirical, suggest that KOM++ can offer significant benefits in efficiency and robustness. Using KOM with a $C_0$-universal kernel such as Gaussian or Matérn is non-parametric, just like optimal matching, and guarantees consistency regardless of model specification, which is particularly reassuring in an observational study. At the same time, the empirical results provide strong evidence that, when applied correctly using KOM++ with ARD for parameter tuning, using these nonparametric kernels incurs little to no deterioration in efficiency compared to using parametric kernels like quadratic when quadratic happens to be well-specified (which of course could not be known in practice). Therefore, a robust general-purpose recom-mendation for the use of KOM in practice is to use the Gaussian or Matérn kernel with KOM++ with ARD for parameter tuning. The choice of which $C_0$-universal kernel seems to matter little. In theory, the Matérn kernel only requires the existence of enough derivatives for "correct specification" and a speed up from $o_p(1)$ to $O_p(1/\sqrt{n})$. The Gaussian kernel requires more but it is more standard in practice and is simpler in form. In practice, such abstract notations of specification may have little relevance as both kernels yield very similar MSEs and what matters most is the reassuring blanket guarantee of model-free consistency, which is shared by both.

As a matching method, KOM is amenable to the same inference methods, such as that of [11], which was developed for optimal matching and generalized in [15, §19.6]. It also gives rise to new inference methods based on the empirical Bayesian estimation of hyperparameter used in KOM++. These enable the construction of confidence intervals for KOM++ estimates.

More importantly, KOM provides an explicit bound on bias, which can be useful in instances with limited overlap where the bias can be significant or even ultimately irreducible. It is again advisable to use a $C_0$-universal kernel for constructing interval estimates using KOM as in eq. (4.2). The question for balance is whether the matched samples are comparable for the purpose of effect estimation. This means that evaluating balance just on differences of covariates means (as for example done on a Love plot) will not be helpful if effects are nonlinear. Evaluating balance using KOM with a nonparametric, $C_0$-universal kernel, however, will necessarily protect against any possible form of the effect. It will also be smaller than a similar bound for a pair-matching method because the imbalances it will leave will necessarily be huge, especially for data with more than just a couple dimensions. Correspondingly, the interval estimate produced by KOM with a $C_0$-universal kernel will be both useful and reliable in practice even when overlap is low and therefore both estimating effects and assessing specification is difficult.

**5. Existing Matching Methods as GOM.**   As mentioned in Sec. 3.5, many matching methods commonly used in practice – not just NNM and 1:1M – are also GOM. Like NNM and 1:1M, many are GOM with $\lambda = 0$. By automatically selecting $\lambda$ using hyperparameter estimation, we extend these methods and reduce estimation error.

5.1. *Nearest-Neighbor Matching.*   Thm. 3 established that NNM is equivalent to $\mathrm{GOM}(\mathcal{W}^{\mathrm{simplex}}, \|\cdot\|_{\mathrm{Lip}(\delta)}, 0)$. To further reduce CMSE, we should consider accounting for the variance due to weighting. We define BVENNM as $\mathrm{GOM}(\mathcal{W}^{\mathrm{simplex}}, \|\cdot\|_{\mathrm{Lip}(\delta)}, \lambda)$, which is given by the following linearly-constrained convex-quadratic optimization problem:

$$
\begin{aligned}
\min \quad & \left(\textstyle\sum_{i\in\mathcal{T}_0, j\in\mathcal{T}_1} \delta(X_i, X_j) S_{ij}\right)^2 + \lambda \left\|W\right\|_2^2 \\
\text{s.t.} \quad & W \in \mathbb{R}^{\mathcal{T}_0}, \; S \in \mathbb{R}_+^{\mathcal{T}_0 \times \mathcal{T}_1} \\
& \textstyle\sum_{i\in\mathcal{T}_0} S_{ij} = 1/n_1 && \forall j \in \mathcal{T}_1 \\
& \textstyle\sum_{j\in\mathcal{T}_1} S_{ij} = W_i && \forall i \in \mathcal{T}_0
\end{aligned}
$$

Ex. 1 illustrates the need to carefully weigh the imperative to balance in the face of variance in NNM but also begs the question of how should we appropriately tune the exchange rate $\lambda$. We present a approach we call NNM++ based on the interpretation of optimal matching as protecting against Lipschitz continuous functions and using cross validation for hyperparameter estimation. Assuming homoskedasticity, the hyperparameters of interest are the residual variance $\sigma^2$ and Lipschitz constant $\gamma$.

NNM++ proceed as follows. We consider regularization parameters $\Psi \subseteq [0,\infty)$ and $m$ disjoint folds $\mathcal{T}_0 = \mathcal{T}_0^{(1)} \sqcup \cdots \sqcup \mathcal{T}_0^{(m)}$. For each $\psi \in \Psi$ and validation fold $k = 1, \ldots, m$, we find $\hat{f}_0^{(k)}$ that minimizes the sum of squared errors in $\mathcal{T}_0 \backslash \mathcal{T}_0^{(k)}$ regularized by $\psi$ times the Lipschitz constant. Out of fold, a range of functions agrees with the fitted value $\hat{v}_i = \hat{f}_0^{(k)}(X_i)$, $\hat{\gamma} = \|\hat{f}_0^{(k)}\|_{\mathrm{Lip}(\delta)}$: $\hat{f}_0^{(k)}(x) \in [\min_{i\in\mathcal{T}_0\backslash\mathcal{T}_0^{(k)}} (\hat{v}_i + \hat{\gamma}\delta(X_i, x)), \, \max_{i\in\mathcal{T}_0\backslash\mathcal{T}_0^{(k)}} (\hat{v}_i - \hat{\gamma}\delta(X_i, x))]$. For the purposes of evaluating out-of-fold error, we use the point-wise midpoint of this interval. We select $\hat{\psi}$ with least out-of-fold mean squared error averaged over folds, let $\hat{\sigma}^2$ be this least error, and refit the $\hat{\psi}$ regularized problem on the whole $\mathcal{T}_0$ sample to estimate $\hat{\gamma}$. This cross-validation procedure is summarized in Alg. 1. Note that we are not interested in a good fit of $\hat{f}_0$ – only a handle on the hyperparameter $\lambda = \sigma^2/\gamma^2$. Finally, we compute the BVENNM weights with $\hat{\lambda} = \hat{\sigma}^2/\hat{\gamma}^2$.

Note that given only that $f_0$ is Lipschitz, is not generally possible to estimate its Lipschitz constant without bias given noisy observations. The above cross-validation procedure will necessarily shrink the estimate and have some downward bias. It is also a very computationally intensive procedure, require solving many large quadratic optimization problems. We merely present NNM++ as one principled way to trade off balance and variance in optimal matching that seems to perform adequately. An alternative approach to finding something on the efficient frontier would be to compute 1:1M and find *anything* on the frontier that dominates it, *e.g.*, in Ex. 1, we may choose any point on the frontier that is below or to the left of 1:1M in Fig. 1b and improve upon it.

The behavior of NNM++ in the case of Ex. 1 using 10-fold cross validation is shown in Figs. 1b and 1c. In a strict sense, NNM++ is *not* "honest" because $\hat{\lambda}$ depends on $Y_{\mathcal{T}_0}$. However, as we only extract a single parameter $\hat{\lambda}$, we guard against data mining (as seen by the resulting low CMSE).

5.2. *Optimal-Caliper Matching.* A sibling of NNM is optimal-caliper matching (OCM), which selects matches with distances that all fit in the smallest possible single caliper. When allowing replacement, NNM is always one of many OCM solutions. Without replacement, NNM and OCM differ. The next theorem shows that OCM is GOM with $\lambda = 0$.

THEOREM 17.  *Fix a pseudo-metric $\delta : \mathcal{X} \times \mathcal{X} \to \mathbb{R}_+$. Let*

$$\|f\|_{\partial(\mu,\delta)} = \mathbb{E}_{x \sim \mu, x' \sim \mu}\left[\delta(x,x')^{-1}|f(x) - f(x')| \mid x \neq x'\right].$$

*OCM with replacement is equivalent to $\mathrm{GOM}(\mathcal{W}^{simplex}, \|\cdot\|_{\partial(\hat{\mu}_n,\delta)}, 0)$, where $\hat{\mu}_n$ is the empirical distribution of $X$. OCM without replacement is equivalent to $\mathrm{GOM}(\mathcal{W}^{1/n_1 \text{-}simplex}, \|\cdot\|_{\partial(\hat{\mu}_n,\delta)}, 0)$.*

5.3. *Coarsened Exact Matching.* Given a coarsening function $C : \mathcal{X} \to \{1,\dots,J\}$ stratifying $\mathcal{X}$, CEM [5] minimizes the coarsened $L_1$ distance,

(5.1)      $\sum_{j=1}^{M}\left|\frac{1}{n_1}\sum_{i \in \mathcal{T}_1}\mathbb{I}_{[C(X_i)=j]} - \sum_{i \in \mathcal{T}_0}W_i\mathbb{I}_{[C(X_i)=j]}\right|,$

by simply equating the matched control distribution in each stratum by setting $W_i = n_1^{-1}|\{j \in \mathcal{T}_1 : [C(X_i) = C(X_j)]\}| / |\{j \in \mathcal{T}_0 : [C(X_i) = C(X_j)]\}|$. CEM is also GOM with $\lambda = 0$.

THEOREM 18.  *Fix a coarsening function $C : \mathcal{X} \to \{1,\dots,J\}$. Let*

$$\|f\|_{L_p(C)} = \left\{ \begin{array}{ll} \|(\sup_{x \in \mathcal{C}^{-1}(j)}|f(x)|)_{j=1}^{J}\|_p & |f(C^{-1}(j))| = 1 \; \forall j, \\ \infty & otherwise. \end{array} \right.$$

*(I.e., the vector $p$-norm of the values taken by $f$ on the coarsened regions if $f$ is piecewise constant.) CEM is equivalent to $\mathrm{GOM}(\mathcal{W}^{simplex}, \|\cdot\|_{L_\infty(C)}, 0)$.*

We can also consider Balance-Variance Efficient CEM (BVECEM) given by $\mathrm{GOM}(\mathcal{W}^{\text{simplex}}, \|\cdot\|_{L_\infty(C)}, \lambda)$ for general $\lambda$. The BVECEM weights are given by the following optimization problem
(5.2)
$\min_{W \in \mathcal{W}^{\text{simplex}}} (\sum_{j=1}^{J}|\frac{1}{n_1}\sum_{i \in \mathcal{T}_1}\mathbb{I}_{[C(X_i)=j]} - \sum_{i \in \mathcal{T}_0}W_i\mathbb{I}_{[C(X_i)=j]}|)^2 + \lambda\,\|W\|_2^2$

Unlike CEM, the solution does not have a closed form. We can solve this optimization problem explicitly by considering all combinations of signs for the $J$ absolute values. For each combination, a Lagrange multiplier argument yields an optimal solution. By further observing that we need only consider monotonic deviations from the usual CEM solution, we obtain Alg. 2, which finds the BVECEM weights in $O(J^2)$ time.

We also consider a CEM++ variant given by estimating $\lambda$ and using the corresponding BVECEM. Unlike BVENNM, here the class of functions $\{f : \|f\|_{L_\infty(C)} < \infty\}$ is generally "small."[14] Therefore, we sidestep a complicated validation scheme and simply estimate the parameters in-sample but use the one-standard-error rule [46, §7.10] to carefully tune $\hat\gamma$. Set $\hat\mu_j = \sum_{i \in \mathcal{T}_0} \mathbb{I}_{[C(X_i)=j]} Y_i / \sum_{i \in \mathcal{T}_0} \mathbb{I}_{[C(X_i)=j]}$ and $\hat\sigma^2 = \frac{1}{n_0-J} \sum_{i \in \mathcal{T}_0} (Y_i - \hat\mu_{C(X_i)})^2$. To estimate $\gamma$, we seek the smallest $\hat\gamma \geq 0$ such that the minimal error over functions with range $2\hat\gamma$ is no worse than $\hat\sigma^2$ plus its standard error $\frac{\hat\sigma^2 \sqrt{2}}{n_0-J}$. (Note $\hat\sigma^2$ is achieved by $2\hat\gamma = \max_j \hat\mu_j - \min_j \hat\mu_j$.) That is, we seek smallest $\hat\gamma \geq 0$ with $\frac{1}{n_0-J} \min_{c \in \mathbb{R}} \sum_{i \in \mathcal{T}_0} (Y_i - \max(\min(\hat\mu_i, c+\hat\gamma), c-\hat\gamma))^2 \leq \hat\sigma^2 (1+\sqrt{2}/(n_0-J))$. We do this by using bisection search on $\hat\gamma$ and a nested golden section search on $c$. We refer to BVECEM with $\hat\lambda = \hat\sigma^2/\hat\gamma^2$ as CEM++.

5.4. *Mean matching and near-fine balance.* Suppose $\mathcal{X} \subseteq \mathbb{R}^d$ so $X_i$ is vector-valued. Mean matching [3, 27, 28] are methods that find a subset of control units $\mathcal{T}_0' \subseteq \mathcal{T}_0$ to reduce the Mahalanobis distance between the sample means

$$M_V(\mathcal{T}_0') = \|V^{1/2}(\tfrac{1}{n_1} \sum_{i \in \mathcal{T}_1} X_i - \tfrac{1}{|\mathcal{T}_0'|} \sum_{i \in \mathcal{T}_0'} X_i)\|_2.$$

We can consider optimal mean matching (OMM) as fully minimizing $M_V(\mathcal{T}_0')$ over all possible subsets $\mathcal{T}_0'$, which we show is GOM.

THEOREM 19. *Suppose $\mathcal{X} \subseteq \mathbb{R}^d$. Let $V \in \mathbb{R}^{d \times d}$ be positive definite and*

$$\|f\|_{2\text{-}lin(V)}^2 = \begin{cases} \alpha^2 + \beta^T V^{-1} \beta & f(x) = \alpha + \beta^T x, \\ \infty & \text{otherwise.} \end{cases}$$

*Then OMM is equivalent to* $\mathrm{GOM}(\mathcal{W}^{subsets}, \|\cdot\|_{2\text{-}lin(V)}, 0)$.

Again, we may consider the more general $\mathrm{GOM}(\mathcal{W}^{\text{subsets}}, \|\cdot\|_{2\text{-}\mathrm{lin}(V)}, \lambda)$. As per Thm. 5 this can be written as the following convex-quadratic binary optimization problem for some $n(\lambda)$ and with $U_i = n(\lambda)W_i$:

$$\min_{U \in \{0,1\}^{\mathcal{T}_0} : \sum_{i \in \mathcal{T}_0} U_i = n(\lambda)} (\sum_{i \in \mathcal{T}_1} \tfrac{X_i}{n_1} - \sum_{i \in \mathcal{T}_0} \tfrac{U_i X_i}{n(\lambda)})^T V (\sum_{i \in \mathcal{T}_1} \tfrac{X_i}{n_1} - \sum_{i \in \mathcal{T}_0} \tfrac{U_i X_i}{n(\lambda)}).$$

---

[14] Specifically, parametric with fewer parameters than data, $J < n_0$. In NNM++, the class of functions was not only infinite dimensional but also *non-separable*!

An alternative form of mean matching would be to minimize the $\ell_p$ distance between sample means. If we define $\|x \mapsto \alpha + \beta^T x\|_{p\text{-lin}} = \|(\alpha, \beta)\|_p$ (and $\infty$ for all non-linear functions) then $\mathrm{GOM}(\mathcal{W}^{\mathrm{subsets}}, \|\cdot\|_{p\text{-lin}}, \lambda)$ is given by the following convex binary optimization problem for some $n(\lambda)$

$$(5.3) \qquad \min_{U \in \{0,1\}^{\mathcal{T}_0} : \sum_{i \in \mathcal{T}_0} U_i = n(\lambda)} \|\sum_{i \in \mathcal{T}_1} \frac{X_i}{n_1} - \sum_{i \in \mathcal{T}_0} \frac{U_i X_i}{n(\lambda)}\|_{p'},$$

where $1/p + 1/p' = 1$. This optimization problem is an integer *linear* optimization problem whenever $p' \in \{1, \infty\}$ (equivalently, $p \in \{1, \infty\}$).

If the covariates are 0-1 indicators (*e.g.*, if they are 2-level factors, if multilevel and encoded in unary as concatenated one-hot vectors, or if continuous and coarsened into multilevel factors and thus encoded), then the sample mean is simply the vector of sample proportions, *i.e.*, it is all marginal distributions of each multilevel factor. In this specific case, mean matching (especially when $p' = 1$, or equivalently $p = \infty$) is known as *near*-fine balance [3]. In this case, we refer to $\mathrm{GOM}(\mathcal{W}^{\mathrm{subsets}}, \|\cdot\|_{\infty\text{-lin}}, 0)$ as optimal near-fine balance (ONFB). When the marginal sample distributions can be made exactly equal, the resulting allocation is known as fine balance [29].

We can also consider a variant given by estimating $\lambda$ for automatic subset size selection. Assuming $d < n_0$, the function class is "small," so we estimate $\lambda$ in-sample. We let $\hat{\alpha}, \hat{\beta}, \hat{\sigma}$ be given by OLS regression on $\{(X_i, Y_i) : i \in \mathcal{T}_0\}$ and set $\hat{\lambda} = \hat{\sigma}^2 / \|\hat{\beta}\|_p^2$. We refer to $\mathrm{GOM}(\mathcal{W}^{\mathrm{subsets}}, \|\cdot\|_{\infty\text{-lin}}, \hat{\lambda})$ as ONFB++.

5.5. *Mixed objectives.*  Methods such as [7, 3] seek to minimize both the sum of pairwise distances and a discrepancy in means. These are also GOM.

THEOREM 20.   *Let* $\mathcal{W}$, $\|\cdot\|_A$, $\|\cdot\|_B$, *and* $\rho > 0$ *be given. Then*

$$\mathfrak{B}(W; \|\cdot\|_{\|\cdot\|_A \oplus_\rho \|\cdot\|_B}) = \mathfrak{B}(W; \|\cdot\|_A) + \rho\,\mathfrak{B}(W; \|\cdot\|_B)$$
$$\text{where} \quad \|f\|_{\|\cdot\|_A \oplus_\rho \|\cdot\|_B} = \inf_{f_A + f_B = f} \max\left\{\|f_A\|_A, \|f_B\|_B / \rho\right\}$$

Therefore, for example, $\mathrm{GOM}(\mathcal{W}^{\mathrm{subsets}}, \|\cdot\|_{\|\cdot\|_{\mathrm{Lip}(\delta)} \oplus_\rho \|\cdot\|_{\infty\text{-lin}}}, \lambda)$ is given by the following integer linear optimization problem (for some $n(\lambda)$):

$$
\begin{aligned}
\min \quad & \frac{1}{n(\lambda)} \sum_{i \in \mathcal{T}_0, j \in \mathcal{T}_1} \delta(X_i, X_j) S_{ij} + \frac{\rho}{n(\lambda)} \sum_{k=1}^{d} D_k \\
\text{s.t.} \quad & U \in \{0,1\}^{\mathcal{T}_0}, S \in \mathbb{R}_+^{\mathcal{T}_0 \times \mathcal{T}_1}, D \in \mathbb{R}^d \\
& \sum_{i \in \mathcal{T}_0} U_i = n(\lambda) \\
& \sum_{i \in \mathcal{T}_0} S_{ij} = n(\lambda)/n_1 && \forall j \in \mathcal{T}_1 \\
& \sum_{j \in \mathcal{T}_1} S_{ij} = U_i && \forall i \in \mathcal{T}_1 \\
& D_k \geq \frac{n(\lambda)}{n_1} \sum_{i \in \mathcal{T}_1} X_{ik} - \sum_{i \in \mathcal{T}_0} U_i X_{ik} && \forall k = 1, \ldots, d \\
& D_k \geq \sum_{i \in \mathcal{T}_0} U_i X_{ik} - \frac{n(\lambda)}{n_1} \sum_{i \in \mathcal{T}_1} X_{ik} && \forall k = 1, \ldots, d
\end{aligned}
$$

### Fig 7: The Balance-Variance Trade-off in CEM and NFB

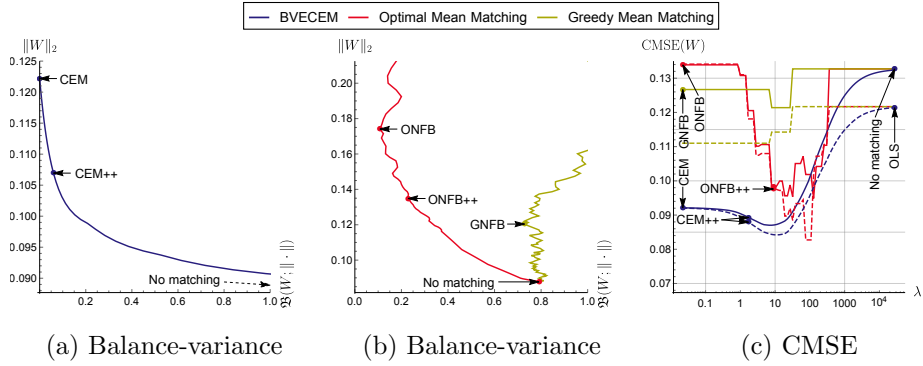

(a) Balance-variance      (b) Balance-variance      (c) CMSE

For $n(\lambda)$ chosen a priori, this is one of the problems considered by [3]. Note our formulation has only $n_0$ discrete variables as we need not constrain $S$ to be integral because the matching polytope is totally unimodular [50].

Similarly, given a coarsening function $C : \mathcal{X} \to \{1, \ldots, J\}$, [7, eq. (2)] is given by $\text{GOM}(\mathcal{W}^{n_1\text{-subset}}, \|\cdot\|_{\|\cdot\|_{\text{Lip}(\delta)} \oplus \rho \|\cdot\|_{L_\infty(C)}}, 0)$, [7, eq. (3)] is given by $\text{GOM}(\mathcal{W}^{n_1\text{-subset}}, \|\cdot\|_{\|\cdot\|_{\text{Lip}(\delta)} \oplus \rho \|\cdot\|_{L_1(C)}}, 0)$, and [7, eq. (4)] is given by $\text{GOM}(\mathcal{W}^{n_1\text{-subset}}, \|\cdot\|_{\|\cdot\|_{\text{Lip}(\delta)} \oplus \rho \|\cdot\|_{L_2(C)}}, 0)$,[15] all for $\rho > 0$ sufficiently large.

EXAMPLE 6. Let us revisit Ex. 1 to study coarsened exact and near-fine balance matching. We coarsen each of the two covariates into an 8-level factor encoding its marginal octile in the sample.[16] We consider BVECEM with all $J = 64$ strata and plot the achievable balance-variance landscape in Fig. 7a. We point out both CEM and CEM++. No matching is in the far right, outside the plot area. We also consider mean-matching ($p = \infty, p' = 1$) on the resulting 16-dimensional vector, which corresponds to near-fine balance on the two coarsened covariates, and plot the balance-variance landscape in Fig. 7b. The red curve in Fig. 7b is the balance-variance achieved by $\text{GOM}(\mathcal{W}^{n_0'\text{-subset}}, \|\cdot\|_{\infty\text{-lin}}, 0)$ for $n_0' \in \{1, \ldots, n_0\}$ ($n_0' \leq 22$ is outside the plot area). ONFB is the leftmost point and has $n_0' = 33$. Note that not all points on the curve are on the efficient frontier given by $\text{GOM}(\mathcal{W}^{\text{subsets}}, \|\cdot\|_{\infty\text{-lin}}, \lambda)$ for $\lambda \in [0, \infty]$. In particular, any $n_0' < 33$ cannot be on the frontier (showing the converse of Thm. 5 is false). We also point out ONFB++ and no matching in the plot. The yellow curve in Fig. 7b is a commonly used greedy heuristic for mean-matching whereby, starting from an empty subset, we incrementally add the unused control unit that would minimize the mean-matching objective. We point out GNFB, given by choosing the point along the greedy path with minimal mean-matching objective. It is the leftmost point on the curve. We plot the resulting CMSE of $\hat{\tau}_W$ (solid) and $\hat{\tau}_{W,\hat{f}_0}$ (dashed) in Fig. 7c corresponding only to points on the efficient frontier of each curve. The need to tune $\lambda$ and consider the variance objective is clear. Even GNFB beats ONFB because the unintended larger matched set induced by sub-optimality. Correctly tuning $\lambda$, ONFB++ improves on both. Similarly, CEM++ improves on CEM.

---

[15]Using the vector 2-norm scaled by $(\sum_{i \in \mathcal{T}_1} \mathbb{I}_{[C(X_i)=j]})_{j=1}^{J}$.

[16]In particular, we chose the greatest number $\ell$ of levels such that for the resulting $\ell^2$ strata, every stratum with a treated unit had at least one control unit.

5.6. *Regression as GOM.* A very popular alternative to matching is regression adjustment via OLS with interaction terms [8]:[17]

$$\hat{\tau}_{\mathrm{OLS}} = \operatorname*{argmin}_{\tau \in \mathbb{R}} \min_{\alpha \in \mathbb{R}, \beta_1, \beta_2 \in \mathbb{R}^d} \sum_{i=1}^{n} \left( Y_i - \alpha - \tau T_i - \beta_1^T X_i - \beta_2^T (X_i - \overline{X}_{\mathcal{T}_1}) T_i \right)^2,$$

where $\overline{X}_{\mathcal{T}_1} = \frac{1}{n_1} \sum_{i \in \mathcal{T}_1} X_i$ is the treated sample mean vector. Surprisingly, this is exactly equivalent to an *unrestricted* version of mean-matching.

THEOREM 21. *Let $V$ positive definite be given and let $W$ be given by* $\mathrm{GOM}(\mathcal{W}^{general}, \| \cdot \|_{2\text{-}lin(V)}, 0)$. *Then $\hat{\tau}_W = \hat{\tau}_{OLS}$.*

We actually prove this as a corollary of a more general result about the ridge-regression version of the regression adjustment with interaction terms:

$$\hat{\tau}_{\lambda\text{-ridge}} = \operatorname*{argmin}_{\tau \in \mathbb{R}} \min_{\alpha \in \mathbb{R}, \beta_1, \beta_2 \in \mathbb{R}^d} \sum_{i=1}^{n} \left( Y_i - \alpha - \tau T_i - \beta_1^T X_i - \beta_2^T (X_i - \overline{X}_{\mathcal{T}_1}) T_i \right)^2 \\ + \lambda \| \beta_1 \|_2^2 + \lambda \alpha^2.$$

THEOREM 22. *Let $W$ be given by* $\mathrm{GOM}(\mathcal{W}^{general}, \| \cdot \|_{2\text{-}lin}, \lambda)$ *for $\lambda \geq 0$. Then $\hat{\tau}_W = \hat{\tau}_{\lambda\text{-ridge}}$.*

These results reveal a very close connection between matching and regression adjustment and expands the scope of existing connections [51, 38]. But, there are several nuanced but important differences between regression adjustment and regular mean-matching (*i.e.*, with simplex or subset weights). Matching with simplex or subset weights results in a distribution over the sample units and thus is more interpretable, preserving the unit of analysis. This also allows certain randomization-based inferences. Moreover, whereas linear regression is subject to dangerous extrapolation [52, 4], matching with weights in the simplex (corresponding to convex combinations), including subsets, inherently prohibits extrapolation. OLS and mean-matching may coincide only if *exact* fine balance is feasible.

---

[17]Note that using the interaction term $(X_i - \overline{X}_{\mathcal{T}_1})T_i$ corresponds to estimating the effect on the treated whereas using the interaction term $(X_i - \overline{X})T_i$, as it appears in [8], corresponds to estimating the overall average effect.

**6. Conclusion.** In this paper, we presented an encompassing framework and theory for matching methods for causal inference, arising from generalizations of a new functional analytical interpretation of optimal matching. On the one hand, this framework revealed a unifying thread between and provided a unified theoretical analysis for a variety of commonly used methods, including both matching and regression methods. This in turn lead to new extensions to methods subsumed in this framework that appropriately and automatically adjust the balance-variance trade-off inherent in matching revealed by the theory developed. These extensions lead to benefits in estimation error relative to their standard counterparts.

On the other hand, this framework lead to the development of a new class of matching methods based on kernels. The new methods, called KOM, were shown to have some of the more appealing properties of the different methods in common use, as supported by specialized theory developed. In particular, KOM yields either a distribution over or a subset of the control units preserving the unit of analysis and avoiding extrapolation, KOM has favorable consistency properties yielding parametric-rate estimation under correct specification and model-free consistency regardless thereof, KOM has favorable robustness and efficiency properties when used in an augmented weighted estimator, KOM has similarly favorable robustness properties when used as preprocessing before regression, KOM++ judiciously and automatically weighs balance in the face of variance, and KOM allows for flexible model selection via empirical Bayes methods. These properties make KOM a particularly apt tool for causal inference.

**References.**

---

**ALGORITHM 1:** Cross-Validation Estimation for NNM++

---

**input:** Control data $Y_{\mathcal{T}_0}$, distance matrix $D \in \mathbb{R}^{\mathcal{T}_0 \times \mathcal{T}_0}$, regularizer grid $\Psi \subseteq \mathbb{R}_+$, and number of folds $m$.

Randomly split the control data into disjoint folds $\mathcal{T}_0 = \mathcal{T}_0^{(1)} \sqcup \cdots \sqcup \mathcal{T}_0^{(m)}$.

**for** $k \in \{1, \ldots, m\}$, $\psi \in \Psi$: **do**

   Solve $\quad \begin{aligned} &\min_{\hat{v}, \hat{\gamma}} \quad \frac{1}{|i \in \mathcal{T}_0 \backslash \mathcal{T}_0^{(k)}|} \sum_{i \in \mathcal{T}_0 \backslash \mathcal{T}_0^{(k)}} (\hat{v}_i - y_i)^2 + \psi \hat{\gamma} \\ &\text{s.t.} \quad \hat{v}_i - \hat{v}_j \le \hat{\gamma} D_{ij} \quad \forall i, j \in \mathcal{T}_0 \backslash \mathcal{T}_0^{(k)} \end{aligned}$ .

   Set $\hat{Y}_i = \frac{1}{2}(\min_{j \in \mathcal{T}_0 \backslash \mathcal{T}_0^{(k)}} (\hat{v}_j + \hat{\gamma} D_{ij}) + \max_{j \in \mathcal{T}_0 \backslash \mathcal{T}_0^{(k)}} (\hat{v}_j - \hat{\gamma} D_{ij}))$.

   Set $\hat{\sigma}_{k,\psi}^2 = \sum_{i \in \mathcal{T}_0^{(k)}} (Y_i - \hat{Y}_i)^2 / (|\mathcal{T}_0^{(k)}| - 1)$.

**end for**

Set $\hat{\sigma}_{\psi}^2 = \frac{1}{m} \sum_{k=1}^{m} \hat{\sigma}_{k,\psi}^2$ for $\psi \in \Psi$, $\hat{\psi} = \operatorname{argmin}_{\psi \in \Psi} \hat{\sigma}_{\psi}^2$ and $\hat{\sigma}^2 = \min_{\psi \in \Psi} \hat{\sigma}_{\psi}^2$.

Solve $\quad \begin{aligned} &\min_{\hat{v}, \hat{\gamma}} \quad \frac{1}{|\mathcal{T}_0|} \sum_{i \in \mathcal{T}_0} (\hat{v}_i - y_i)^2 + \hat{\psi} \hat{\gamma} \\ &\text{s.t.} \quad \hat{v}_i - \hat{v}_j \le \hat{\gamma} D_{ij} \quad \forall i, j \in \mathcal{T}_0 \end{aligned}$ .

**output:** $\hat{\lambda} = \hat{\sigma}^2 / \hat{\gamma}^2$.

---

**ALGORITHM 2:** BVECEM (solves eq. (5.2))

---

**input:** Data $X_{1:n}$, $T_{1:n}$, coarsening function $C : \mathcal{X} \to \{1, \ldots, J\}$, and exchange $\lambda$.

Let $n_{tj} = \sum_{i \in \mathcal{T}_t} \mathbb{I}_{[C(X_i) = j]}$ for $t = 0, 1, j = 1, \ldots, J$.

Let $q_j = n_{1j} / (n_1 n_{0j})$ and sort $q_{j_1} \le \cdots \le q_{j_J}$ $(q_{j_0} = -\infty, q_{j_{J+1}} = \infty)$.

Set $v^* = \infty$.

**for** $J_+ = 0, \ldots, J$, $J_- = 0, \ldots, J - J_+$ **do**

   Set $\begin{aligned} n_{0+} &= \sum_{k=1}^{J_+} n_{0j_{J+1-k}}, \quad n_{0-} = \sum_{k=1}^{J_-} n_{0j_k}, \quad r = \sum_{k=J_-+1}^{J+1-J_+} n_{0j_k} q_{j_k}, \\ r_{\Delta} &= \sum_{k=1}^{J_+} n_{0j_{J+1-k}} q_{0j_{J+1-k}} - \sum_{k=1}^{J_-} n_{0j_k} q_{j_k}, \quad r_2 = \sum_{k=J_-+1}^{J-J_+} n_{0j_k} q_{j_k}^2, \\ w_+ &= \frac{2n_{0-}(1-r+r_{\Delta}) + \lambda(1-r)}{4n_{0+}n_{0-} + \lambda(n_{0+} + n_{0-})}, \quad w_- = \frac{2n_{0+}(1-r-r_{\Delta}) + \lambda(1-r)}{4n_{0+}n_{0-} + \lambda(n_{0+} + n_{0-})}, \text{ and} \\ v &= (\sum_{k=1}^{J_+} n_{0j_{J+1-k}} |q_{j_{J+1-k}} - w_+| + \sum_{k=1}^{J_-} n_{0j_k} |q_{j_k} - w_-|)^2 \\ &\quad + \lambda(n_{0p} w_+^2 + n_{0m} w_-^2 + r_2). \end{aligned}$

   **if** $v < v^*$ **then**

      Set $W_i = \begin{cases} 1/n_1 & i \in \mathcal{T}_1, \\ w_- & i \in \mathcal{T}_0, q_{C(X_i)} \le q_{j_{J_-}}, \\ w_+ & i \in \mathcal{T}_0, q_{C(X_i)} \ge q_{j_{J+1-J_+}}, \\ q_{C(X_i)} & \text{otherwise.} \end{cases}$

   **end if**

**end for**

**output:** $W$.

---

## APPENDIX A: CONNECTIONS TO AND GENERALIZATION OF EQUAL-PERCENT BIAS REDUCTION

Equal-percent bias reduction (EPBR) [53, 54] is a property of matching methods stipulating that, on average, the reduction in discrepancy in the mean vector be equal across all the the covariates relative to doing no matching at all. This is the case if and only if the matching reduces imbalance for any linear function of the covariates. In this section we discuss connections to KOM as well as non-linear generalizations of EPBR.

For the sake of exposition, we define EPBR somewhat differently and explain the connection below.

DEFINITION 6. For $\mathcal{X} \subseteq \mathbb{R}^d$, a matching method $W$ is *linearly EPBR* relative to $W'$, written $W \preceq_{\text{lin-EPBR}} W'$, if $|\mathbb{E}[B(W; f)]| \leq |\mathbb{E}[B(W'; f)]|$ for any $f(x) = \beta^T x + \alpha$, $\beta \in \mathbb{R}^d$, $\alpha \in \mathbb{R}$.

Usually, we compare relative to the no-matching weights $W_i^{(0)} = 1/n_0$. The definition of EPBR in [53] is equivalent to saying in our definition that $W$ is comparable to $W^{(0)}$ in lin-EPBR, *i.e.*, either $W \preceq_{\text{lin-EPBR}} W^{(0)}$ or $W^{(0)} \preceq_{\text{lin-EPBR}} W$. This equivalence is proven in the following. The definition in [53] is stated in terms of the equivalent proportionality statement below, relative to $W^{(0)}$ and without magnitude restrictions on $\alpha$.

THEOREM 23. *Suppose $\mathcal{X} \subseteq \mathbb{R}^d$. Then, $W \preceq_{\text{lin-EPBR}} W'$ if and only if $\mathbb{E}[\frac{1}{n_1} \sum_{i \in \mathcal{T}_1} X_i - \sum_{i \in \mathcal{T}_0} W_i X_i] = \alpha \mathbb{E}[\frac{1}{n_1} \sum_{i \in \mathcal{T}_1} X_i - \sum_{i \in \mathcal{T}_0} W_i' X_j]$ for some $\alpha \in [-1, 1]$ as vectors in $\mathbb{R}^d$.*

In [55], the authors show that in the special case of proportionally ellipsoidal distributions, affinely invariant matching methods are EPBR. We restate and reprove the result in terms of our definitions.

DEFINITION 7. For $\mathcal{X} \subseteq \mathbb{R}^d$, a matching method $W$ is *affinely invariant* if $W(X_{1:n}, T_{1:n}) = W(X_{1:n}A^T + \mathbf{1}_n a^T, T_{1:n})$ for all non-singular $A \in \mathbb{R}^{d \times d}$ and $a \in \mathbb{R}^d$ (*i.e.*, $x \mapsto Ax + a$ is applied to each data point separately).

DEFINITION 8. Two random vectors $Z$ and $Z'$ are *proportionally ellipsoidal* if there exists PSD matrix $\Sigma$, $\alpha, \alpha' \in \mathbb{R}_+$, and a characteristic function $\phi : \mathbb{R} \to \mathbb{C}$ such that, for any $v$, $v^T Z$ and $v^T Z'$ have characteristic functions $e^{iv^T \mathbb{E}[Z]t} \phi(\alpha v^T \Sigma v t^2)$ and $e^{iv^T \mathbb{E}[Z']t} \phi(\alpha' v^T \Sigma v t^2)$, respectively.

THEOREM 24. *If $X \mid T = 0$, $X \mid T = 1$ are proportionally ellipsoidal and $W$ is affinely invariant, then $W$ and $W^{(0)}$ are lin-EPBR comparable.*

---

**ALGORITHM 3:** Affine Invariance by Studentization

---

**input:** Data $X_{1:n}, T_{1:n}$, a matching method $W(X_{1:n}, T_{1:n})$.

Let $\hat{\mu} = \frac{1}{n_0} \sum_{i \in \mathcal{T}_0} X_i$, $\widehat{\Sigma} = \frac{1}{n_0 - 1} \sum_{i \in \mathcal{T}_0} (X_i - \hat{\mu})(X_i - \hat{\mu})^T$.

Eigen-decompose $\widehat{\Sigma} = U \operatorname{Diag}(\tau_1, \ldots, \tau_d) U^T$.

Set $\widehat{\Sigma}^{\dagger/2} = U \operatorname{Diag}(\mathbb{I}_{[\tau_1 \neq 0]} \tau_1^{-1/2}, \ldots, \mathbb{I}_{[\tau_d \neq 0]} \tau_d^{-1/2}) U^T$.

**output:** $W((X_{1:n} - \hat{\mu})\widehat{\Sigma}^{\dagger/2}, T_{1:n})$.

---

Methods based on the Mahalanobis distance are affinely invariant. KOM with a fitted scaling matrix is also affinely invariant as discussed in 4.6. All unitarily invariant matching methods can be made to be affinely invariant by preprocessing the data. This procedure is detailed in Alg. 3. KOM is unitarily invariant if the kernel is unitarily invariant, including all kernels studied in this paper. With the exception of CEM, all methods we have studied have been unitarily invariant.

DEFINITION 9.   For $\mathcal{X} \subseteq \mathbb{R}^d$, a matching method $W$ is *unitarily invariant* if $W(X_{1:n}, T_{1:n}) = W(X_{1:n} A^T, T_{1:n})$ for all unitary $A = A^{-T} \in \mathbb{R}^{d \times d}$. Equivalently, $W$ is *unitarily invariant* if it only depends on $X_{1:n} X_{1:n}^T, T_{1:n}$.

THEOREM 25.   *Suppose $\mathcal{X} \subseteq \mathbb{R}^d$. If $W$ is unitarily invariant then Alg. 3 produces an affinely invariant weighting method.*

At the same time, affinely invariant methods only make sense if we have more datapoints than the dimension of the data, since the dimension of the data is precisely the dimension of the space of linear functions on the data.

THEOREM 26.   *Suppose $\mathcal{X} \subseteq \mathbb{R}^d$. If $W$ is affinely invariant then $W$ is constant over all data $X_{1:n}$ that is affinely independent.*[18]

In other words, if $n \leq d$, then any affinely invariant matching method does nothing useful at all because it is (generically) invariant to the data.

The above exposition establishes the connection of KOM to EPBR and the benefits it bestows. If we either fit a scaling matrix by marginal likelihood or studentize the data, then KOM is affinely invariant and hence EPBR for proportionally ellipsoidal data, *i.e.*, has uniform improvement over all linear outcomes. However, one of the main attractions of KOM is in dealing with *non-linear* outcomes. We next present a direct generalization of EPBR to non-linear outcomes, allowing us to characterize when matching methods can have performance guarantees over families of *non-linear* outcomes. We recreate the analogous lin-EPBR results for the non-linear version.

---

[18]$X_{1:n}$ are said to be affinely independent if $X_{2:n} - X_1$ are linearly independent.

DEFINITION 10.    A matching method $W$ is $\mathcal{F}$-*EPBR* relative to $W'$, written $W \preceq_{\mathcal{F}\text{-EPBR}} W'$, if $|\mathbb{E}\left[B(W; f)\right]| \leq |\mathbb{E}\left[B(W'; f)\right]|$ for every $f \in \mathcal{F}$.

THEOREM 27.    *Let $\mathcal{F}$ be a linear subspace of the functions $\mathcal{X} \to \mathbb{R}$ under pointwise addition and scaling. Then, $W \preceq_{\mathcal{F}\text{-EPBR}} W'$ if and only if there exists $\alpha \in [-1, 1]$ such that, as operators on $\mathcal{F}$, $\mathbb{E}\left[B(W; \cdot)\right] = \alpha \mathbb{E}\left[B(W'; \cdot)\right]$.*

DEFINITION 11.    Let $\mathcal{K}$ be a PSD kernel on $\mathcal{X}$ and let $\mathcal{F}$ be its RKHS. The $\mathcal{X}$-valued random variables $Z$ and $Z'$ are *proportionally $\mathcal{K}$-ellipsoidal* if there exist $\mu, \mu' \in \mathcal{F}$, PSD compact $C \in \mathcal{F} \otimes \mathcal{F}$, $\alpha, \alpha' \in \mathbb{R}_+$, and a characteristic function $\phi : \mathbb{R} \to \mathbb{C}$ such that, for every $x \in \mathcal{X}$, $K(Z, x)$ and $K(Z', x)$ are a real random variables distributed with characteristic functions $e^{i\mu(x)t}\phi(C(x,x)t^2)$, $e^{i\mu'(x)t}\phi(\alpha C(x,x)t^2)$ respectively.

For example, proportionally ellipsoidal is equivalent to proportionally $\mathcal{K}$-ellipsoidal with $\mathcal{K}(x, x') = x^T x'$.

DEFINITION 12.    Let $\mathcal{K}$ be a PSD kernel on $\mathcal{X}$ and let $\mathcal{F}$ be its RKHS. A matching method $W$ is *$\mathcal{K}$-affinely invariant* if $\exists W' : \mathcal{F}^n \times \{0, 1\}^n \to \mathcal{W}$ such that for any bounded non-singular $A \in \mathcal{F} \otimes \mathcal{F}$ and $a \in \mathcal{F}$, we have $W(X_{1:n}, T_{1:n}) = W'(\{A\mathcal{K}(X_i, \cdot) + a\}_{i=1}^n, T_{1:n})$.

THEOREM 28.    *Let $\mathcal{K}$ be a PSD kernel on $\mathcal{X}$ and let $\mathcal{F}$ be its RKHS. If $X \mid T = 0$, $X \mid T = 1$ are proportionally $\mathcal{K}$-ellipsoidal and $W$ is affinely invariant then $W$ is $\mathcal{F}$-EPBR relative to $W^{(0)}$.*

DEFINITION 13.    Let $\mathcal{K} : \mathcal{X} \times \mathcal{X} \to \mathbb{R}$ be a PSD kernel. $W$ is *$\mathcal{K}$-unitarily invariant* if it depends on the data via its Gram matrix, *i.e.*, $\exists W' : \mathcal{S}_+^{n \times n} \times \{0, 1\}^n \to \mathcal{W}$ such that $W(X_{1:n}, T_{1:n}) = W'((\mathcal{K}(X_i, X_j))_{i,j=1}^n, T)$.

For example, KOM is $\mathcal{K}$-unitarily invariant.

THEOREM 29.    *If $W'$ is $\mathcal{K}$-unitarily invariant then Alg. 4 produces a $\mathcal{K}$-affinely invariant weighting method.*

However, there are limits to $\mathcal{K}$-affine invariance. The following shows that $\mathcal{K}$-affine invariance only makes sense for non-universal kernels since all $C_0$-universal kernels are strictly positive definite (has Gram matrix that is positive definite whenever all datapoints are distinct).

---

**ALGORITHM 4:** $\mathcal{K}$-Affine Invariance by $\mathcal{K}$-Studentization

---

**input:** Data $X_{1:n}, T_{1:n}$, a PSD kernel $\mathcal{K}$, a $\mathcal{K}$-unitarily invariant $W(X, T)$.
Let $K_{ij} = \mathcal{K}(X_i, X_j)$, $E_{ij}^{(0)} = \mathbb{1}_{[j \in \mathcal{T}_0]}/n_0$.
Set $K^C = (I - E^{(0)}) K (I - E^{(0)})^T$, $M_{ij} = \sum_{l \in \mathcal{T}_0} K_{il}^C K_{ji}^C$.
Compute the pseudo-inverse $M^\dagger$ and let $\overline{K} = K^C M^\dagger K^C$.
**output:** $W'(\overline{K}, T_{1:n})$.

---

THEOREM 30. *Suppose $\mathcal{K}$ is strictly positive definite and $W$ is $\mathcal{K}$-affinely invariant. Then, $W$ is constant over all $X_{1:n}$ that are distinct.*

## APPENDIX B: PROOFS

PROOF OF THM. 1. Define

$$(B.1) \qquad \Xi(W) = \tfrac{1}{n_1} \sum_{i \in \mathcal{T}_1} \epsilon_i - \sum_{i \in \mathcal{T}_0} W_i \epsilon_i.$$

Rewrite SATT $= \frac{1}{n_1} \sum_{i \in \mathcal{T}_1} Y_i - \frac{1}{n_1} \sum_{i \in \mathcal{T}_1} Y_i(0)$, it is clear SATT differs from $\hat{\tau}_W$ only in the second term so that, letting $W_i = 1/n_1$ for $i \in \mathcal{T}_1$,

$$\hat{\tau} - \text{SATT} = \tfrac{1}{n_1} \sum_{i \in \mathcal{T}_1} Y_i(0) - \sum_{i \in \mathcal{T}_0} W_i Y_i(0) = B(W; f_0) + \Xi(W)$$
$$= \sum_{i=1}^{n} (-1)^{T_i+1} W_i f_0(X_i) + \sum_{i=1}^{n} (-1)^{T_i+1} W_i \epsilon_i,$$

For each $i$, we have

$$\mathbb{E}[\epsilon_i \mid X_{1:n}, T_{1:n}] = \mathbb{E}[Y_i(0) \mid X_i, T_i] - f_0(X_i) = \mathbb{E}[Y_i(0) \mid X_i] - f_0(X_i) = 0,$$

where the first equality is by definition of $\epsilon_i$ and the second by Asn. 1. Since $W_i = W_i(X_{1:n}, T_{1:n})$, we have $\mathbb{E}[(-1)^{T_i+1} W_i \epsilon_i \mid X_{1:n}, T_{1:n}] = 0$. For each $i, j$,

$$\mathbb{E}\left[(-1)^{T_i+T_j} W_i W_j \epsilon_i \epsilon_j \mid X_{1:n}, T_{1:n}\right] = (-1)^{T_i+T_j} W_i W_j \mathbb{E}[\epsilon_i \epsilon_j \mid X_{1:n}, T_{1:n}]$$
$$= W_i W_j \operatorname{Cov}(Y_i(0), Y_j(0)) = \mathbb{1}_{[i=j]} W_i^2 \sigma_i^2,$$
$$\mathbb{E}[(-1)^{T_i+T_j} W_i W_j \epsilon_i f_0(X_j) \mid X_{1:n}, T_{1:n}] = 0,$$

completing the proof. $\qquad\square$

PROOF OF THM. 3. Let $D$ be the distance matrix $D_{ii'} = \delta(X_i, X_{i'})$. By the definition of the Lipschitz norm and linear optimization duality we get,

$$\mathfrak{B}(W; \|\cdot\|_{\text{Lip}(\delta)}) = \sup_{v_i - v_{i'} \leq D_{ii'}} \tfrac{1}{n_1} \sum_{i \in \mathcal{T}_1} v_1 - \sum_{i \in \mathcal{T}_0} W_i v_i$$

$$= \min_{S \in \mathbb{R}_+^{n \times n}} \quad \sum_{i,i'} D_{ii'} S_{ii'}$$
$$\text{s.t.} \quad \sum_{i'=1}^{n} (S_{ii'} - S_{i'i}) = 1/n_1 \quad \forall i \in \mathcal{T}_1$$
$$\sum_{i'=1}^{n} (S_{ii'} - S_{i'i}) = -W_i \quad \forall i \in \mathcal{T}_0.$$

This describes a min-cost flow problem with sources $\mathcal{T}_1$ with inputs $1/n_1$, sinks $\mathcal{T}_0$ with outputs $W_i$, edges between every two nodes with costs $D_{ii'}$ and without capacities. Consider any source $i \in \mathcal{T}_1$ and any sink $i' \in \mathcal{T}_0$ and any path $i, i_1, \ldots, i_m, i'$. By the triangle inequality, $D_{ii'} \leq D_{ii_1} + D_{i_1 i_2} + \cdots + D_{i_m i'}$. Therefore, as there are no capacities, it is always preferable to send the flow from the sources to the sinks along the direct edges from $\mathcal{T}_1$ to $\mathcal{T}_0$. That is, we can eliminate all other edges and write

$$\mathfrak{B}(W; \|\cdot\|_{\mathrm{Lip}(\delta)}) = \min_{S \in \mathbb{R}_+^{\mathcal{T}_1 \times \mathcal{T}_0}} \quad \sum_{i \in \mathcal{T}_1, \, i' \in \mathcal{T}_0} D_{ii'} S_{ii'}$$
$$\text{s.t.} \quad \sum_{i' \in \mathcal{T}_0} S_{ii'} = 1/n_1 \quad \forall i \in \mathcal{T}_1$$
$$\sum_{i' \in \mathcal{T}_1} S_{i'i} = W_i \qquad \forall i \in \mathcal{T}_0.$$

For the case of NNM, using the transformation $W_i' = n_1 W_i$, we get

$$\min_{W \in \mathcal{W}^{\mathrm{simplex}}} \mathfrak{B}(W; \|\cdot\|_{\mathrm{Lip}(\delta)}) = \frac{1}{n_1} \min_{S, W'} \quad \sum_{i \in \mathcal{T}_1, \, i' \in \mathcal{T}_0} D_{ii'} S_{ii'}$$
$$\text{s.t.} \quad S \in \mathbb{R}_+^{\mathcal{T}_1 \times \mathcal{T}_0}, \; W' \in \mathbb{R}_+^{\mathcal{T}_0}$$
$$\sum_{i \in \mathcal{T}_0} W_i' = n_1$$
$$\sum_{i' \in \mathcal{T}_0} S_{ii'} = 1 \qquad \forall i \in \mathcal{T}_1$$
$$\sum_{i \in \mathcal{T}_1} S_{ii'} - W_{i'}' = 0 \qquad \forall i' \in \mathcal{T}_0.$$

This describes a min-cost netwrok flow problem with sources $\mathcal{T}_1$ with inputs 1; nodes $\mathcal{T}_0$ with 0 exogenous flow; one sink with output $n_1$; edges from each $i \in \mathcal{T}_1$ to each $i' \in \mathcal{T}_0$ with flow variable $S_{ii'}$, cost $D_{ii'}$, and without capacity; and edges from each $i \in \mathcal{T}_0$ to the sink with flow variable $W_i'$ and without cost or capacity. Because all data is integer, the optimal solution of is integer (see [50]). This solution (in terms of $W'$) is equal to sending the whole input 1 from each source in $\mathcal{T}_1$ to the node in $\mathcal{T}_0$ with smallest distance and from there routing this flow to the sink, which corresponds exactly to NNM.

For the case of 1:1M, using the same transformation, we get

$$\min_{W \in \mathcal{W}^{1/n_1 \text{-simplex}}} \mathfrak{B}(W; \|\cdot\|_{\mathrm{Lip}(\delta)}) = \frac{1}{n_1} \min_{S, W'} \quad \sum_{i \in \mathcal{T}_1, \, i' \in \mathcal{T}_0} D_{ii'} S_{ii'}$$
$$\text{s.t.} \quad S \in \mathbb{R}_+^{\mathcal{T}_1 \times \mathcal{T}_0}, \; W' \in \mathbb{R}_+^{\mathcal{T}_0}$$
$$\sum_{i \in \mathcal{T}_0} W_i' = n_1$$
$$W_i' \leq 1 \quad \forall i \in \mathcal{T}_0$$
$$\sum_{i' \in \mathcal{T}_0} S_{ii'} = 1 \qquad \forall i \in \mathcal{T}_1$$
$$\sum_{i \in \mathcal{T}_1} S_{ii'} - W_i' = 0 \quad \forall i' \in \mathcal{T}_0.$$

This describes the same min-cost netwrok flow problem except that the edges from each $i \in \mathcal{T}_0$ to the sink have a capacity of 1. Because all data is integer, the optimal solution is integer. Since the optimal $S_{ii'}$ is integer, by $\sum_{i' \in \mathcal{T}_0} S_{ii'} = 1$, for each $i \in \mathcal{T}_1$ there is exactly one $i' \in \mathcal{T}_0$ with $S_{ii'} = 1$ and all others are zero. $S_{ii'} = 1$ denotes matching $i$ with $i'$. The optimal $W'_i$ is integral and so, by $W'_i \leq 1$, $W'_i \in \{0, 1\}$. Hence, for each $i \in \mathcal{T}_0$, $\sum_{i' \in \mathcal{T}_1} S_{ii'} \in \{0, 1\}$ so we only use node $i$ at most once. The cost of $S$ is exactly the sum of pairwise distances in the match. Hence, the optimal solution corresponds exactly to 1:1M. $\square$

PROOF OF COR. 4. Let $W$ be given by $\mathrm{GOM}(\mathcal{W}, \|\cdot\|, \lambda)$. Then we have that $V^2(W; \sigma^2_{1:n}) \leq \sigma^2(\|W\|_2 + 1/n_1) = \gamma^2 \lambda(\|W\|_2 + 1/n_1)$ and $B^2(W; f_0) \leq \inf_{g:B(W;g)=0} \forall_{W \in \mathcal{W}} B^2(W; f_0 + g) \leq \gamma \mathfrak{B}(W; \|\cdot\|)$. Finally, apply Thm. 1. $\square$

PROOF OF THM. 5. Let $W$ be given by $\mathrm{GOM}(\mathcal{W}^{\text{subsets}}, \|\cdot\|, \lambda)$ and let $n(\lambda) = \|W\|_0 = |\{i \in \mathcal{T}_0 : W^*_i > 0\}|$. Then $W \in \mathcal{W}^{n(\lambda)\text{-subset}}$ so $W$ is also given by $\mathrm{GOM}(\mathcal{W}^{n(\lambda)\text{-subset}}, \|\cdot\|, \lambda)$. However, $\|W'\|_2^2 = 1/n(\lambda)$ for all $W' \in \mathcal{W}^{n(\lambda)\text{-subset}}$, so the variance term is constant among this space of weights. Therefore, $W$ is given by $\mathrm{GOM}(\mathcal{W}^{n(\lambda)\text{-subset}}, \|\cdot\|, 0)$. $\square$

PROOF OF THM. 6. Note $\mathrm{argmin}_{W \in \mathcal{W}^{n(\lambda)\text{-multisubset}}} \|W\|_2^2 = \mathcal{W}^{n(\lambda)\text{-subset}}$. So by definition of GOM for $\lambda = \infty$ we get the equivalence between the first and the second. The equivalence between the second and third was argued in the proof of Thm. 5. $\square$

PROOF OF THM. 7. This follows because $\|W\|_2^2$ is convex, $\mathfrak{B}(W; \|\cdot\|)$ is nonnegative and is the supremum of affine functions in $W$ (*i.e.*, $B(\lambda W + (1 - \lambda)W'; f) = \lambda B(W; f) + (1 - \lambda)B(W'; f))$ and hence convex, and the square is convex and nondecreasing on the nonnegative line. $\square$

PROOF OF THM. 8. Define $F(W) = \sup_{\|f\| \leq 1} \sum_{i=1}^n (2T_i - 1)W_i f(X_i)$ for $W \in \mathbb{R}^n$, and rewrite problem (3.1) redundantly as

$$\min \ r + \lambda s$$
(B.2) $$\text{s.t. } W \in \mathbb{R}^n, \ r \in \mathbb{R}, s \in \mathbb{R}, t \in \mathbb{R}$$
(B.3) $$W_{\mathcal{T}_0} \in \mathcal{W}^{\text{simplex}}$$
(B.4) $$W_{\mathcal{T}_1} = e_{n_1}/n_1$$
(B.5) $$\|W\|_2^2 \leq s$$
(B.6) $$t^2 \leq r$$
(B.7) $$F(W) \leq t, \ \|W\|_2 \leq 1$$

Since each of (B.2)-(B.7) are convex, by [56, Thm. 4.2.2] we reduce the $\epsilon$-optimization problem to $\epsilon$-separation on each of (B.2)-(B.7), which is immediately trivial for (B.2)-(B.6) in polynomial time in $n$. By [56, Thms. 4.2.5, 4.2.7] we reduce $\epsilon$-separation on (B.7) to $\epsilon$-violation, which by binary search reduces $\epsilon$-violation on $S_t = \{W \in \mathbb{R}^{TC} : F(W) \leq v, \|W\|_2 \leq 1\}$ for fixed $t \geq 0$. If $t = 0$, then $S_t = \{0\}$ and we are done. Otherwise, letting $E_i : f \mapsto f(X_i)$ and $M_i = \||E_i|\|_* < \infty$, note that $F(W)$ is continuous since

$$
\begin{aligned}
F(W) - F(W') &= \sup_{\|f\| \leq 1} B(W_{\mathcal{T}_0}; f) - \sup_{\|f\| \leq 1} B(W'_{\mathcal{T}_0}; f) \\
&\leq \sup_{\|f\| \leq 1} B(W_{\mathcal{T}_0} - W'_{\mathcal{T}_0}; f) \\
&\leq \sup_{\|f\| \leq 1} \|W_{\mathcal{T}_0} - W'_{\mathcal{T}_0}\|_2 (\textstyle\sum_{i=1}^n f(X_i))^{1/2} \\
&\leq \|W_{\mathcal{T}_0} - W'_{\mathcal{T}_0}\|_2 \|M\|_2
\end{aligned}
$$

Therefore, since $F(0) = 0$, we have $\{\|W\|_2 \leq t/\|M\|_2\} \subset S_t \subset \{\|W\|_2 \leq 1\}$. Since we can check membership by evaluating $\|W\|_2$ and $F(w)$, by [56, Thm. 4.3.2] we have an $\epsilon$-violation algorithm. $\qquad\square$

PROOF OF THM. 17. Using similar arguments to the proof of Thm. 3, we get that $\mathfrak{B}(W; \|\cdot\|_{\partial(\hat{\mu}_n, \delta)})$ is equal (up to a scaling of $n(n-1)$) to

$$
\begin{aligned}
\min_{S \in \mathbb{R}_+^{\mathcal{T}_1 \times \mathcal{T}_0}, t \in \mathbb{R}_+} \quad & t \\
\text{s.t.} \quad & \textstyle\sum_{i' \in \mathcal{T}_0} S_{ii'} = 1/n_1 \quad \forall i \in \mathcal{T}_1 \\
& \textstyle\sum_{i' \in \mathcal{T}_1} S_{i'i} = W_i \quad \forall i \in \mathcal{T}_0 \\
& t \geq D_{ii'} S_{ii'} \qquad\qquad \forall i \in \mathcal{T}_1, i' \in \mathcal{T}_0.
\end{aligned}
$$

Hence, minimizing it over $\mathcal{W}^{\text{simplex}}$ or $\mathcal{W}^{1/n_1\text{-simplex}}$ we get the same network flow problems except with a bottleneck objective. The solution is still integer and gives the pair matching with minimal maximal pair distance, corresponding exactly to OCM with or without replacement. $\qquad\square$

PROOF OF THM. 18. For $f$ piecewise constant on the coarsening components let $f_j$ denote its value on the $j^{\text{th}}$ component. By writing $f(x) = \sum_{j=1}^M \mathbb{I}_{[C(x)=j]} f_j$ and exchanging sums we can rewrite $B(W; f)$ as

$$
B(W; f) = \textstyle\sum_{j=1}^J f_j \left(\frac{1}{n_1} \sum_{i \in \mathcal{T}_1} \mathbb{I}_{[C(X_i)=j]} - \sum_{i \in \mathcal{T}_0} W_i \mathbb{I}_{[C(X_i)=j]}\right).
$$

Under the constraint $|f_j| \leq 1 \ \forall j$, the maximizer of the above assigns $\pm 1$ to each $f_j$ in order to make the $j^{\text{th}}$ term nonnegative. Hence,

$$
\mathfrak{B}(W; \|\cdot\|_{L_\infty(C)}) = \textstyle\sum_{j=1}^J |\frac{1}{n_1} \sum_{i \in \mathcal{T}_1} \mathbb{I}_{[C(X_i)=j]} - \sum_{i \in \mathcal{T}_0} W_i \mathbb{I}_{[C(X_i)=j]}|,
$$

which we recognize as the coarsened $L_1$ distance from eq. (5.1). $\qquad\square$

PROOF OF THM. 19. Let $W \in \mathcal{W}^{\text{subsets}}$ and $\mathcal{T}_0' = \{i \in \mathcal{T}_0 : W_i > 0\}$. Then, by duality of Euclidean norms,

$$\mathfrak{B}(W; \|\cdot\|_{2\text{-lin}(V)}) = \sup_{\beta^T V \beta \leq 1} \beta^T (\tfrac{1}{n_1} \sum_{i \in \mathcal{T}_1} X_i - \sum_{i \in \mathcal{T}_0} W_i X_i) = M_V(\mathcal{T}_0'). \quad \square$$

PROOF OF THM. 20. By linearity, we have

$$\begin{aligned}
\mathfrak{B}(W; \|\cdot\|_{\|\cdot\|_{\mathrm{A}} \oplus_\rho \|\cdot\|_{\mathrm{B}}}) &= \sup_{\|f\|_{\|\cdot\|_{\mathrm{A}} \oplus_\rho \|\cdot\|_{\mathrm{B}}} \leq 1} B(W; f) \\
&= \sup_{\|f_A\|_{\mathrm{A}} \leq 1, \|f_B\|_{\mathrm{B}}/\rho \leq 1} B(W; f_A + f_B) \\
&= \mathfrak{B}(W; \|\cdot\|_{\mathrm{A}}) + \rho \mathfrak{B}(W; \|\cdot\|_{\mathrm{B}}) \qquad\qquad \square
\end{aligned}$$

PROOF OF THM. 21. Note that $\text{GOM}(\mathcal{W}^{\text{general}}, \|\cdot\|_{2\text{-lin}(V)}, 0)$ is the same as $\text{GOM}(\mathcal{W}^{\text{general}}, \|\cdot\|_{2\text{-lin}}, 0)$ applied to the data $\tilde{X}_i = V^{1/2} X_i$. However, applying OLS adjustment to data $X_i$ or data $\tilde{X}_i$ is exactly the same because $\beta_1, \beta_2$ are unrestricted so we can make the transformation $\tilde{\beta}_1, \tilde{\beta}_2 = V^{-1/2} \beta_1, V^{-1/2} \beta_2$ without any effect except transforming $\tilde{X}_i$ to $X_i$. Finally, we apply Thm. 22 with $\lambda = 0$. $\square$

PROOF OF THM. 22. We can rewrite the ridge-regression problem as

$$\begin{aligned}
&\min_{\tau, \alpha, \beta_1, \beta_2} \sum_{i=1}^n (Y_i - \alpha - \tau T_i - \beta_1^T X_i - \beta_2^T (X_i - \overline{X}_{\mathcal{T}_1}) T_i)^2 + \lambda \|\beta_1\|_2^2 + \lambda \alpha^2 \\
=&\min_{\tau, \alpha, \beta_1, \beta_2} \bigg( \sum_{i \in \mathcal{T}_0} (Y_i - \alpha - \beta_1^T X_i)^2 + \lambda \|\beta_1\|_2^2 + \lambda \alpha^2 \\
&\qquad\qquad\qquad + \sum_{i \in \mathcal{T}_1} (Y_i - (\alpha + \tau - \beta_2^T \overline{X}_{\mathcal{T}_1}) - (\beta_1 + \beta_2)^T X_i)^2 \bigg) \\
=&\min_{\alpha, \beta_1} (\sum_{i \in \mathcal{T}_0} (Y_i - \alpha - \beta_1^T X_i)^2 + \lambda \|\beta_1\|_2^2 + \lambda \alpha^2) + \min_{\tilde{\alpha}, \tilde{\beta}} \sum_{i \in \mathcal{T}_1} (Y_i - \tilde{\alpha} - \tilde{\beta}^T X_i)^2,
\end{aligned}$$

where we used the transformation $\tilde{\alpha} = \alpha + \tau - \beta_2^T \overline{X}_{\mathcal{T}_1}$, $\tilde{\beta} = \beta_1 + \beta_2$ and the fact that $\tau$ and $\beta_2$ are unrestricted to see that $\tilde{\alpha}, \tilde{\beta}$ are unrestricted. Because $\tilde{\alpha}, \tilde{\beta}$ solve an OLS problem with intercept on $\mathcal{T}_1$, the mean of in-sample residuals are zero, and therefore:

$$\begin{aligned}
0 &= \tfrac{1}{n_1} \sum_{i \in \mathcal{T}_1} (Y_i - \tilde{\alpha} - \tilde{\beta}^T X_i) = \overline{Y}_{\mathcal{T}_1}(1) - \tilde{\alpha} - \beta_1^T \overline{X}_{\mathcal{T}_1} - \beta_2^T \overline{X}_{\mathcal{T}_1}, \\
&\implies \tau_{\lambda\text{-ridge}} = \tilde{\alpha} - \alpha + \beta_2^T \overline{X}_{\mathcal{T}_1} = \overline{Y}_{\mathcal{T}_1}(1) - \alpha - \beta_1^T \overline{X}_{\mathcal{T}_1} \\
&\qquad\qquad = \overline{Y}_{\mathcal{T}_1}(1) - (\alpha, \beta_1^T) \tilde{X}_{\mathcal{T}_1}^T \tfrac{e_{n_1}}{n_1},
\end{aligned}$$

where $\tilde{X}_i = (1, X_i)$ and $\tilde{X}_{\mathcal{T}_t}$ is the $n_t \times (d+1)$ matrix of these. Note $\alpha, \beta_1$ solve ridge-regression on $\mathcal{T}_0$, so:

$$(\alpha, \beta_1^T)^T = (\tilde{X}_{\mathcal{T}_0}^T \tilde{X}_{\mathcal{T}_0} + \lambda I_{d+1})^{-1} \tilde{X}_{\mathcal{T}_0}^T Y_{\mathcal{T}_0}.$$

Therefore, letting $\tilde{W} = \frac{1}{n_1}\tilde{X}_{\mathcal{T}_0}(\tilde{X}_{\mathcal{T}_0}^T\tilde{X}_{\mathcal{T}_0} + \lambda I_{d+1})^{-1}\tilde{X}_{\mathcal{T}_1}^T e_{n_1}$, we must have

$$\tau_{\lambda\text{-ridge}} = \overline{Y}_{\mathcal{T}_1}(1) - \sum_{i \in \mathcal{T}_0}\tilde{W}_i Y_i.$$

On the other hand, the weights $W$ given by $\mathrm{GOM}(\mathcal{W}^{\text{general}}, \|\cdot\|_{2\text{-lin}}, \lambda)$ minimize the following objective function:

$$\sup_{\alpha^2 + \|\beta\|_2^2 \leq 1}\left(\frac{1}{n_1}\sum_{i \in \mathcal{T}_1}(\alpha + \beta^T X_i) - \sum_{i \in \mathcal{T}_0}W_i(\alpha + \beta^T X_i)\right)^2 + \lambda\|W\|_2^2$$

$$= \sup_{\alpha^2 + \|\beta\|_2^2 \leq 1}\left(\left(\begin{array}{c}W\\-\frac{e_{n_1}}{n_1}\end{array}\right)^T\tilde{X}\left(\begin{array}{c}\alpha\\\beta\end{array}\right)\right)^2 + \lambda\|W\|_2^2 = \left\|\tilde{X}^T\left(\begin{array}{c}W\\-\frac{e_{n_1}}{n_1}\end{array}\right)\right\|_2^2 + \lambda\|W\|_2^2$$

$$= \|\tilde{X}_{\mathcal{T}_0}^T W - \tilde{X}_{\mathcal{T}_1}^T\frac{e_{n_1}}{n_1}\|_2^2 + \lambda\|W\|_2^2.$$

By first order optimality conditions on unrestricted $W$ we have

$$0 = \tilde{X}_{\mathcal{T}_0}(\tilde{X}_{\mathcal{T}_0}^T W - \tilde{X}_{\mathcal{T}_1}^T\frac{e_{n_1}}{n_1}) + \lambda W \implies W = (\tilde{X}_{\mathcal{T}_0}\tilde{X}_{\mathcal{T}_0}^T + \lambda I_{n_0})^{-1}\tilde{X}_{\mathcal{T}_0}\tilde{X}_{\mathcal{T}_1}^T\frac{e_{n_1}}{n_1}.$$

Fix $\tilde{\lambda} > 0$. By applying the Sherman-Morrison-Woodbury (SMW) formula [57, §C.4.3] *twice*, we have

$$(\tilde{X}_{\mathcal{T}_0}\tilde{X}_{\mathcal{T}_0}^T + \tilde{\lambda}I_{n_0})^{-1}\tilde{X}_{\mathcal{T}_0} \overset{\text{SMW}}{=} (\frac{1}{\tilde{\lambda}}I_{n_0} - \tilde{X}_{\mathcal{T}_0}(\frac{1}{\tilde{\lambda}}\tilde{X}_{\mathcal{T}_0}^T\tilde{X}_{\mathcal{T}_0} + I_{d+1})^{-1}\tilde{X}_{\mathcal{T}_0}^T)\tilde{X}_{\mathcal{T}_0}$$

$$= \frac{1}{\tilde{\lambda}}\tilde{X}_{\mathcal{T}_0}(I_{d+1} - (\frac{1}{\tilde{\lambda}}\tilde{X}_{\mathcal{T}_0}^T\tilde{X}_{\mathcal{T}_0} + I_{d+1})^{-1}\frac{1}{\tilde{\lambda}}\tilde{X}_{\mathcal{T}_0}^T\tilde{X}_{\mathcal{T}_0})$$

$$\overset{\text{SMW}}{=} \frac{1}{\tilde{\lambda}}\tilde{X}_{\mathcal{T}_0}(\frac{1}{\tilde{\lambda}}\tilde{X}_{\mathcal{T}_0}^T\tilde{X}_{\mathcal{T}_0} + I_{d+1})^{-1}$$

$$= \tilde{X}_{\mathcal{T}_0}(\tilde{X}_{\mathcal{T}_0}^T\tilde{X}_{\mathcal{T}_0} + \tilde{\lambda}I_{d+1})^{-1}.$$

If $\lambda > 0$ then set $\tilde{\lambda} = \lambda$. If $\lambda = 0$ and $\tilde{X}_{\mathcal{T}_0}^T\tilde{X}_{\mathcal{T}_0}$ is invertible then, by continuous transformation over the limit $\tilde{\lambda} \to 0$, the equation of the first to the last holds with $\tilde{\lambda} = 0$. In either case, this shows $W = \tilde{W}$, completing the proof. $\qquad\square$

Proof of Thm. 9. Let $p(x) = \mathbb{P}(T = 1 \mid X = x)$, $p_0 = \mathbb{P}(T = 0)$, $p_1 = \mathbb{P}(T = 1)$. By Asn. 2, there exists $\alpha > 0$ such that $q(X) = \alpha p(X)/(1 - p(X))$ is a.s. in $(0, 1)$. For each $i$, let $\tilde{W}'_i \in \{0, 1\}$ be 1 if $T_i = 1$ and otherwise Bernoulli with probability $q(X_i)$ (where the draw for $i$ is fixed over $n$). Then we have that $X_i \mid T_i = 0, \tilde{W}'_i = 1$ is distributed identically as $X_i \mid T_i = 1$. Let $n'_0 = \sum_{i \in \mathcal{T}_0}\tilde{W}'_i$ and for each $i \in \mathcal{T}_0$ set $\tilde{W}_i = \tilde{W}'_i/n'_0$. For $i \in \mathcal{T}_1$, set $\tilde{W}_i = 1/n_1$. Let $X_i^{(1)}$ be new, independent replicates distributed as $X_i^{(1)} \sim (X \mid T = 1)$. Let $\xi_i(f) = \tilde{W}'_i(f(X_i) - f(X_i^{(1)}))$. Let $A_0 = \frac{1}{n_0}\sum_{i \in \mathcal{T}_0}\xi_i$ and $A_1 = \frac{1}{n_1}\sum_{i \in \mathcal{T}_1}\xi_i$. Then, we have that $B(\tilde{W}; f) = A_1(f) - (n_0/n'_0)A_0(f)$.

By construction of $\tilde{W}'_i$ and $\xi$, we see that $\mathbb{E}\left[\xi_i\right] = 0$ (*i.e.*, Bochner integral). By (vi), $\|\xi\|_*$ has second moment. Also, each $\xi_i$ is independent. Therefore, by [58] (and since $B$-convexity of $\mathcal{F}$ implies $B$-convexity of $\mathcal{F}^*$ [59]), a law of large numbers holds yielding, a.s., $\|A_0\|_* \to 0$ and $\|A_1\|_* \to 0$. Since $(\bar{n}_0/n'_0) \to \alpha/p_1 < \infty$ a.s., we have that $\mathfrak{B}(\tilde{W}; \|\cdot\|) = \|B(\tilde{W}; \cdot)\|_* \to 0$ a.s. Moreover, $\|\tilde{W}_{\mathcal{T}_0}\|_2^2 = 1/n'_0 \to 0$ a.s. since $n'_0 \to \infty$ a.s. Since $\tilde{W}_{\mathcal{T}_0}$ is feasible by (iv) and $W$ is optimal,

$$\mathfrak{E}^2(W; \|\cdot\|, \lambda_n) \le \mathfrak{E}^2(\tilde{W}_{\mathcal{T}_0}; \|\cdot\|, \lambda_n) \le \mathfrak{B}^2(\tilde{W}_{\mathcal{T}_0}; \mathcal{F}) + \bar{\lambda}\|\tilde{W}_{\mathcal{T}_0}\|_2^2 \to 0 \text{ a.s.}$$

By (viii), $\exists\gamma : \|[f_0]\| \le \gamma < \infty$. By (vii), $\exists\sigma^2 : \sigma_i^2 \le \sigma^2$. By Thm. 1,

$$\text{CMSE}(\hat{\tau}_W) \le (\gamma^2 + \sigma^2/\underline{\lambda})\mathfrak{E}^2(W; \|\cdot\|, \lambda_n) + \sigma^2/n_1 \to 0 \text{ a.s.}$$

By iterated expectation, $\hat{\tau}_W - \text{SATT} \to 0$ in $L_2$ and hence in probability. $\quad\square$

PROOF OF THM. 10. Recalling $\Xi(W)$ from eq. (B.1), we can write

$$\hat{\tau}_{W,\hat{f}_0} - \text{SATT} = B(W; \tilde{f}_0 - \hat{f}_0) + B(W; f_0 - \tilde{f}_0) + \Xi(W),$$

From the proof of Thm. 9, we have $\mathfrak{B}(W; \|\cdot\|) = o_p(1)$, $\|W\|_2 = o_p(1)$, $\Xi^2(W) = o_p(1)$. Moreover,

$$|B(W; \tilde{f}_0 - \hat{f}_0)| \le (\|W\|_2^2 + 1/n_1)^{1/2}(\sum_{i=1}^n (\tilde{f}_0(X_i) - \hat{f}_0(X_i))^2)^{1/2}$$
$$= o_p(1)O_p(1) = o_p(1).$$

In case (a), $B(W; f_0 - \tilde{f}_0) = 0$ yields the result. In case (b), $\mathbb{E}|B(W; f_0 - \tilde{f}_0)| \le (\|f_0\| + \|\tilde{f}_0\|)\mathbb{E}\mathfrak{B}(W; \mathcal{K}) = O(n^{-1/2})$, yielding the result. In case (c), we write $\hat{\tau}_{W,\hat{f}_0} - \text{SATT} = B(W; f_0 - \hat{f}_0) + \Xi(W)$ and $|B(W; f_0 - \hat{f}_0)| \le (\|[f_0]\| + \|[\hat{f}_0]\|)\mathfrak{B}(W; \mathcal{K}) = O_p(1)o_p(1) = o_p(1)$. $\quad\square$

PROOF OF THM. 11. Writing $W'_i = T_i/n_1 - (1-T_i)W_i$, by the representer property of $\mathcal{K}$ and by self-duality of Hilbert spaces,

$$\mathfrak{B}^2(W; \|\cdot\|) = \max_{\|f\|\le 1}\left(\sum_{i=1}^n (-1)^{T_i+1}W'_i\left\langle\mathcal{K}(X_i, \cdot), f\right\rangle\right)^2$$
$$= \left\|\sum_{i=1}^n (-1)^{T_i+1}W'_i\mathcal{K}(X_i, \cdot)\right\|^2$$
$$= \left\langle\sum_{i=1}^n (-1)^{T_i+1}W'_i\mathcal{K}(X_i, \cdot), \sum_{i=1}^n (-1)^{T_i+1}W'_i\mathcal{K}(X_i, \cdot)\right\rangle$$
$$= \sum_{i,j=1}^n (-1)^{T_i+T_j}W'_iW'_jK_{ij},$$

which when written in block form gives rise to the result. $\quad\square$

PROOF OF THM. 12. By Thm. 1, we have

$$\mathbb{E}\left[(\hat{\tau}_W - \text{SATT})^2 \mid X, T, f_0\right] = B^2(W; f_0) + \sigma^2 \|W\|_2^2 + \frac{\sigma^2}{n_1}.$$

Marginalizing over $f_0$ and writing $W_i' = T_i/n_1 + (1 - T_i)W_i$, we get

$$
\begin{aligned}
\text{CMSE}(\hat{\tau}_W) &= \sum_{i,j=1}^n (-1)^{T_i + T_j} W_i' W_j' \mathbb{E}\left[f_0(X_i)f_0(X_j) \mid X_{1:n}, T_{1:n}\right] + \sigma \|W'\|_2^2 \\
&= \sum_{i,j=1}^n (-1)^{T_i + T_j} W_i' W_j' \gamma^2 \mathcal{K}(X_i, X_j) + \sigma^2 \|W'\|_2^2 \\
&= \gamma^2 \mathfrak{B}(W; \mathcal{K}) + \sigma^2 \|W'\|_2^2 = \gamma^2 (\mathfrak{E}(W; \mathcal{K}, \lambda) + \lambda/n_1),
\end{aligned}
$$

where the second equality is by the of Gaussian process prior and the third equality is due to Thm. 11. □

For the next proof we use the following lemma.

LEMMA 31. *For random variables $Z_n \geq 0$ and any sub-sigma algebra $\mathcal{G}$, $\mathbb{E}[Z_n \mid \mathcal{G}] = O_p(1) \implies Z_n = O_p(1)$ and $\mathbb{E}[Z_n \mid \mathcal{G}] = o_p(1) \implies Z_n = o_p(1)$.*

PROOF. Suppose $\mathbb{E}[Z_n \mid \mathcal{G}] = O_p(1)$. Let $\nu > 0$ be given. Then $\mathbb{E}[Z_n \mid \mathcal{G}] = O_p(1)$ says that there exist $N, M$ such that $\mathbb{P}(\mathbb{E}[Z_n \mid \mathcal{G}] > M) \leq \nu/2$ for all $n \geq N$. Let $M_0 = \max\{M, 2/\nu\}$ and observe that, for all $n \geq N$,

$$
\begin{aligned}
\mathbb{P}(Z_n > M_0^2) &= \mathbb{P}(Z_n > M_0^2, \mathbb{E}[Z_n \mid \mathcal{G}] > M_0) + \mathbb{P}(Z_n > M_0^2, \mathbb{E}[Z_n \mid \mathcal{G}] \leq M_0) \\
&= \mathbb{P}(Z_n > M_0^2, \mathbb{E}[Z_n \mid \mathcal{G}] > M_0) + \mathbb{E}[\mathbb{P}(Z_n > M_0^2 \mid \mathcal{G})\mathbb{I}_{[\mathbb{E}[Z_n\mid\mathcal{G}]\leq M_0]}] \\
&\leq \nu/2 + \mathbb{E}\left[\frac{\mathbb{E}[Z_n \mid \mathcal{G}]}{M_0^2}\mathbb{I}_{[\mathbb{E}[Z_n\mid\mathcal{G}]\leq M_0]}\right] \leq \nu/2 + 1/M_0 \leq \nu.
\end{aligned}
$$

Now suppose $\mathbb{E}[Z_n \mid \mathcal{G}] = o_p(1)$. Let $\eta > 0, \nu > 0$ be given. Let $N$ be such that $\mathbb{P}(\mathbb{E}[Z_n \mid \mathcal{G}] > \nu\eta/2) \leq \nu/2$. Then for all $n \geq N$:

$$
\begin{aligned}
\mathbb{P}(Z_n > \eta) &= \mathbb{P}(Z_n > \eta, \mathbb{E}[Z_n \mid \mathcal{G}] > \eta\nu/2) + \mathbb{P}(Z_n > \eta, \mathbb{E}[Z_n \mid \mathcal{G}] \leq \eta\nu/2) \\
&= \mathbb{P}(Z_n > \eta, \mathbb{E}[Z_n \mid \mathcal{G}] > \eta\nu/2) + \mathbb{E}[\mathbb{P}(Z_n > \eta \mid \mathcal{G})\mathbb{I}_{[\mathbb{E}[Z_n\mid\mathcal{G}]\leq\eta\nu/2]}] \\
&\leq \nu/2 + \mathbb{E}\left[\frac{\mathbb{E}[Z_n \mid \mathcal{G}]}{\eta}\mathbb{I}_{[\mathbb{E}[Z_n\mid\mathcal{G}]\leq\eta\nu/2]}\right] \leq \nu/2 + \nu/2 \leq \nu. \qquad \square
\end{aligned}
$$

PROOF OF THM. 13. Repeat the proof of Thm. 9 up to defining $\xi_i$. Let $\xi_i'$ an identical and independent replicate of $\xi_i$, conditioned on $T_i$. Let $\rho_i$ be iid Rademacher random variables independent of all else. For $t = 0, 1$,

$$
\begin{aligned}
\mathbb{E}[\|A_t\|_*^2 \mid T_{1:n}] &= \tfrac{1}{n_t^2} \mathbb{E}[\| \textstyle\sum_{i \in \mathcal{T}_t} \xi_i \|_*^2 \mid T_{1:n}] \\
&= \tfrac{1}{n_t^2} \mathbb{E}[\| \textstyle\sum_{i \in \mathcal{T}_t} \left( \mathbb{E}\left[\xi_i'\right] - \xi_i \right) \|_*^2 \mid T_{1:n}] \\
&\leq \tfrac{1}{n_t^2} \mathbb{E}[\| \textstyle\sum_{i \in \mathcal{T}_t} \left( \xi_i' - \xi_i \right) \|_*^2 \mid T_{1:n}] \\
&= \tfrac{1}{n_t^2} \mathbb{E}[\| \textstyle\sum_{i \in \mathcal{T}_t} \rho_i \left( \xi_i' - \xi_i \right) \|_*^2 \mid T_{1:n}] \\
&\leq \tfrac{4}{n_t^2} \mathbb{E}[\| \textstyle\sum_{i \in \mathcal{T}_t} \rho_i \xi_i \|_*^2 \mid T_{1:n}].
\end{aligned}
$$

By (iv), $M = \mathbb{E}[\|\xi_i\|_*^2 \mid T] \leq 4\mathbb{E}[\mathcal{K}(X, X) \mid T = 1] < \infty$. Note that $\|\xi - \zeta\|_*^2 + \|\xi + \zeta\|_*^2 = 2\|\xi\|_*^2 + 2\|\zeta\|_*^2 + 2 \langle \xi, \zeta \rangle - 2 \langle \xi, \zeta \rangle = 2\|\xi\|_*^2 + 2\|\zeta\|_*^2$. By induction, $\sum_{\rho_i \in \{-1, +1\}^m} \| \sum_{i=1}^m \rho_i \xi_i \|_*^2 = 2^m \sum_{i=1}^m \|\xi_i\|_*^2$. Therefore,

$$
\begin{aligned}
\mathbb{E}[\|A_t\|_*^2 \mid T] &\leq \tfrac{4}{n_t^2} \mathbb{E}[\| \textstyle\sum_{i \in \mathcal{T}_t} \rho_i \xi_i \|_*^2 \mid T] \leq \tfrac{4}{n_t^2} \mathbb{E}[\textstyle\sum_{i \in \mathcal{T}_t} \|\xi_i\|_*^2 \mid T] \\
&\leq \tfrac{4}{n_t^2} \textstyle\sum_{i \in \mathcal{T}_t} \mathbb{E}[\|\xi_i\|_*^2 \mid T] = 4M / n_t.
\end{aligned}
$$

Since $\|\tilde{W}_{\mathcal{T}_0}\|_2 = 1 / n_0'$ and $\lambda_n \leq \overline{\lambda}$, we conclude that,

$$
\mathfrak{E}^2(\tilde{W}_{\mathcal{T}_0}; \mathcal{K}, \lambda_n) = \|A_1 + (n_0 / n_0') A_0\|_*^2 + \lambda_n \|\tilde{W}_{\mathcal{T}_0}\|_2^2 = O_p(1/n).
$$

Because $W$ is optimal and by (iii) $\tilde{W}_{\mathcal{T}_0}$ is feasible and $\lambda^{-1} \leq \underline{\lambda}^{-1}$, we then also have $\mathfrak{B}^2(W; \mathcal{K}) = O_p(1/n)$ and $\|W\|_2^2 = O_p(1/n)$. By (v), $\exists \sigma^2 : \sigma_i^2 \leq \sigma^2$. Therefore, $V^2(W; \sigma_{1:n}^2) \leq \sigma^2(\|W\|_2^2 + 1/n_1) = O_p(1/n)$ and hence by Lemma 31 $\Xi(W) = O_p(n^{-1/2})$.

In case (a), we have $B(W; f_0) \leq \|[f_0]\| \mathfrak{B}(W; \mathcal{K}) = O(n^{-1})$. Thm. 1 and Lemma 31 yield the result.

Now consider case (b). Fix $\eta > 0, \nu > 0$. Let $\tau = \sqrt{\nu\eta/3}/M$. Because $\mathfrak{B}(W; \mathcal{K}) = O_p(n^{-1/2}) = o_p(n^{-1/4})$ and $\|W\|_2 = O_p(n^{-1/2})$, there are $M, N$ such that for all $n \geq N$ both $\mathbb{P}(n^{1/4} \mathfrak{B}(W; \mathcal{K}) > \sqrt{\eta}) \leq \nu/3$ and $\mathbb{P}(n^{1/2}(\|W\|_2 + 1/\sqrt{n_1}) > M\sqrt{\eta}) \leq \nu/3$. By existence of second moment, there is $g_0' = \sum_{i=1}^{\ell} \beta_i I_{S_i}$ with $(\mathbb{E}\left[(f_0(X) - g_0'(X))^2\right])^{1/2} \leq \tau/2$ where $I_S(x)$ are the simple functions $I_S(x) = \mathbb{I}[x \in S]$ for $S$ measurable. Let $i = 1, \ldots, \ell$. Let $U_i \supset S_i$ open and $E_i \subseteq S_i$ compact be such that $\mathbb{P}(X \in U_i \backslash E_i) \leq \tau^2/(4\ell |\beta_i|)^2$. By Urysohn's lemma [23], there exists a continuous function $h_i$ with support $C_i \subseteq U_i$ compact, $0 \leq h_i \leq 1$, and $h_i(x) = 1 \, \forall x \in E_i$. Therefore, $(\mathbb{E}\left[(I_{S_i}(X) - h_i)^2\right])^{1/2} = (\mathbb{E}\left[(I_{S_i}(X) - h_i)^2 \mathbb{I}\left[X \in U_i \backslash E_i\right]\right])^{1/2} \leq (\mathbb{P}(X \in U_i \backslash E_i))^{1/2} \leq \tau/(4\ell |\beta_i|)$. By $C_0$-universality, $\exists g_i = \sum_{j=1}^m \alpha_j \mathcal{K}(x_j, \cdot)$ such that $\sup_{x \in \mathcal{X}} |h_i(x) - g_i(x)| < \tau/(4\ell |\beta_i|)$. Because $\mathbb{E}\left[(h_i - g_i)^2\right] \leq \sup_{x \in \mathcal{X}} |h_i(x) - g_i(x)|^2$, we have $\sqrt{\mathbb{E}\left[(I_{S'}(X) - g_i)^2\right]} \leq \tau/(2\ell |\beta_i|)$. Let $g_0 =$

$\sum_{i=1}^{\ell} \beta_i g_i$. Then $(\mathbb{E}\left[(f_0(X) - g_0(X))^2\right])^{1/2} \leq \tau/2 + \sum_{i=1}^{\ell} |\beta_i| \, \tau/(2\ell\,|\beta_i|) = \tau$ and $\|g_0\| < \infty$. Let $\delta_n = \sqrt{\frac{1}{n}\sum_{i=1}^{n}(f_0(X_i) - g_0(X_i))^2}$. Then, we have that

$$
\begin{aligned}
|B(W; f_0)| &\leq |B(W; g_0)| + |B(W; f_0 - g_0)| \\
&\leq \|[g_0]\|\, \mathfrak{B}(W; \mathcal{K}) + \sqrt{n}(\|W\|_2 + 1/\sqrt{n_1})\delta_n.
\end{aligned}
$$

Let $N' = \max\{N, 2\lceil\|[f_0]\|^4/\eta^2\rceil\}$. Then by union bound, for all $n \geq N'$, we have

$$
\begin{aligned}
\mathbb{P}(|B(W; f_0)| > \eta) &\leq \mathbb{P}(n^{-1/4}\|[g_0]\| > \sqrt{\eta}) + \mathbb{P}(n^{1/4}\mathfrak{B}(W; \mathcal{K}) > \sqrt{\eta}) \\
&\quad + \mathbb{P}(n^{1/2}(\|W\|_2 + 1/\sqrt{n_1}) > M\sqrt{\eta}) + \mathbb{P}(\delta_n > \sqrt{\eta}/M) \\
&\leq 0 + \nu/3 + \nu/3 + \nu/3 = \nu.
\end{aligned}
$$

Hence, $|B(W; f_0)| = o_p(1)$. Thm. 1 and Lemma 31 yield the result. $\qquad\square$

PROOF OF THM. 14. From the proof of Thm. 13 we have $\mathfrak{B}^2(W; \mathcal{K}) = O_p(1/n)$, $\|W\|_2^2 = O_p(1/n)$, $\Xi(W) = O_p(n^{-1/2})$. In case (a), we have $|B(W; f_0 - \hat{f}_0)| \leq \|[f_0 - \hat{f}_0]\|\mathfrak{B}(W; \mathcal{K}) = o_p(n^{-1/2})$ and so $\hat{\tau}_{W, \hat{f}_0} - \text{SATT} = \Xi(W) + o_p(n^{-1/2})$, yielding the result. In case (b), we have $|B(W; f_0 - \hat{f}_0)| \leq (\|[f_0]\| + \|[\hat{f}_0]\|)\mathfrak{B}(W; \mathcal{K}) = O_p(1)O_p(n^{-1/2})$, yielding the result.

In the other cases, expand the error of $\hat{\tau}_{W, \hat{f}_0}$:

$$
\begin{aligned}
\hat{\tau}_{W, \hat{f}_0} - \text{SATT} &= B(W; \tilde{f}_0 - \hat{f}_0) + B(W; f_0 - \tilde{f}_0) + \Xi(W). \\
|B(W; \tilde{f}_0 - \hat{f}_0)| &\leq (\|W\|_2^2 + 1/n_1)^{1/2}(\sum_{i=1}^{n}(\tilde{f}_0(X_i) - \hat{f}_0(X_i))^2)^{1/2} \\
&= O_p(n^{-1/2})O_p(1).
\end{aligned}
$$

In case (c), $B(W; f_0 - \tilde{f}_0) = 0$ yields the result. In case (d), we repeat the argument in the proof of Thm. 13 to show that both $|B(W; f_0)| \to 0$ and $|B(W; \tilde{f}_0)| \to 0$, yielding the result. In case (e), $|B(W; f_0 - \hat{f}_0)| \leq (\|f_0\| + \|\hat{f}_0\|)\mathfrak{B}(W; \mathcal{K}) = O_p(n^{-1/2})$, yielding the result. $\qquad\square$

PROOF OF THM. 15. From the proof of Thm. 13, $\mathfrak{B}^2(W; \mathcal{K}) = O(1/n)$, $\|W\|_2^2 = O(1/n)$. From the proof of Thm. 22, for $\hat{f}_0(x) = \hat{\alpha} - \hat{\beta}^T x$ where $\hat{\alpha}, \hat{\beta} = \operatorname{argmin}_{\alpha \in \mathbb{R}, \beta \in \mathbb{R}^d} \sum_{i \in \mathcal{T}_0} W_i(Y_i - \alpha - \beta^T X_i)^2$, we have $\hat{\tau}_{\text{WLS}(W)} = \frac{1}{n_1}\sum_{i \in \mathcal{T}_1}(Y_i - \hat{f}_0(X_i))$. Because least squares with intercept has zero in-sample bias, $\sum_{i \in \mathcal{T}_0} W_i \hat{f}_0(X_i) = \sum_{i \in \mathcal{T}_0} W_i Y_i$, so that by adding and substracting this term, we see that $\hat{\tau}_{\text{WLS}(W)} = \hat{\tau}_{W, \hat{f}_0} = \hat{\tau}_W - B(W; \hat{f}_0)$.

Let $\tilde{X}_i = (1, X_i)$, $\tilde{\beta} = (\hat{\alpha}, \hat{\beta}^T)^T$, $\hat{P} = \tilde{X}_{\mathcal{T}_0}^T W \tilde{X}_{\mathcal{T}_0} = \sum_{i \in \mathcal{T}_0} W_i \tilde{X}_i \tilde{X}_i^T$, $\hat{G} = \sum_{i \in \mathcal{T}_0} W_i f_0(X_i)\tilde{X}_i$, and $\hat{H} = \sum_{i \in \mathcal{T}_0} W_i \tilde{X}_i \epsilon_i$. Then $\tilde{\beta} = \hat{P}^{-1}(\hat{G} + \hat{H})$. Follow

the argument in Thm. 13 for the case of $\mathcal{K}$ being $C_0$-universal to show $\hat{P} \to P$ in probability, where $P = \mathbb{E}[\tilde{X}\tilde{X}^T \mid T = 1]$. By the Schor complement, since $\mathbb{E}[XX^T \mid T = 1]$ is non-singular, $P$ is also non-singular and hence we have $\hat{P}^{-1} \to P^{-1}$ in probability. Follow the argument in Thm. 13 for the case of $\mathcal{K}$ being $C_0$-universal to show $\hat{G} \to G = \mathbb{E}[f_0(X)\tilde{X} \mid T = 1]$ in probability. Moreover, letting $M > 1$ be such that $\sup_{x \in \mathcal{X}} \|x\|_2 \leq M$ by assumption, $\mathbb{E}[\|\hat{H}\|_2^2 \mid X_{1:n}, T_{1:n}] = \sum_{i \in \mathcal{T}_0} W_i^2 \sigma_i^2 \|\tilde{X}_i\|_2^2 \leq 2M^2 \sigma^2 \|W\|_2^2 = O_p(1/n)$ so that by Lemma 31 $\|\hat{H}\|_2^2 = O_p(1/n)$.

Consider case (a). By Thm. 13, $\hat{\tau}_W \to 0$ in probability. By the above, we have $\|\tilde{\beta}\|_2 = O_p(1)$. Let $\eta > 0, \rho > 0$ be given. Then there is $R > 0$ such that $\mathbb{P}(\|\tilde{\beta}\|_2 > R) \leq \rho/3$. Let $M', N'$ be such that $\mathbb{P}(\sqrt{n}(\|W\|_2^2 + 1/n_1)^{1/2} > M') \leq \rho/3$ for all $n \geq N'$. Let $r = \eta/(8MM')$ and $\{\bar{\beta} : \|\bar{\beta} - \bar{\beta}_1\|_2 \leq r\}, \ldots, \{\bar{\beta} : \|\bar{\beta} - \tilde{\beta}_\ell\|_2 \leq r\}$ be a finite cover of the compact $\{\bar{\beta} : \|\bar{\beta}\|_2 \leq R\}$. Let $f_{\bar{\beta}}(x) = \bar{\beta}^T(1, x)$. By $C_0$-universality and boundedness, $\exists g_k$ with $\|g_k\| < \infty$ and $\sup_{x \in \mathcal{X}} |g_k(x) - f_{\bar{\beta}_k}(x)| \leq \eta/(4M')$. Let $\Gamma = \max_{k=1,\ldots,\ell} \|g_k\|$ and let $N'' \geq N'$ be such that $\mathbb{P}(\mathfrak{B}(W; \mathcal{K}) > \eta/(2\Gamma)) \leq \rho/3$ for all $n \geq N''$. Note that $\sup_{x \in \mathcal{X}} |f_{\bar{\beta}}(x) - f_{\bar{\beta}'}(x)| \leq 2M\|\bar{\beta} - \bar{\beta}'\|_2$. Then we have that

$$\sup_{\|\bar{\beta}\| \leq R} |B(W; f_{\bar{\beta}})|$$
$$\leq \sup_{\|\bar{\beta}\| \leq R} \min_k(|B(W; g_k)| + |B(W; f_{\bar{\beta}_k} - g_k)| + |B(W; f_{\bar{\beta}} - f_{\bar{\beta}_k})|)$$
$$\leq \Gamma\mathfrak{B}(W; \mathcal{K}) + \sqrt{n}(\|W\|_2^2 + 1/n_1)^{1/2}(\kappa + 2Mr)$$
$$= \Gamma\mathfrak{B}(W; \mathcal{K}) + \sqrt{n}(\|W\|_2^2 + 1/n_1)^{1/2}\eta/(2M')$$

Finally, for all $n \geq N''$, we have

$$\mathbb{P}(|B(W; \hat{f}_0)| > \eta) \leq \mathbb{P}(\|\tilde{\beta}\| > R) + \mathbb{P}(|B(W; \hat{f}_0)| > \eta, \|\tilde{\beta}\| \leq R)$$
$$\leq \rho/3 + \mathbb{P}(\sup_{\|\bar{\beta}\| \leq R} |B(W; f_{\bar{\beta}})| > \eta)$$
$$\leq \rho/3 + \mathbb{P}(\mathfrak{B}(W; \mathcal{K}) > \eta/(2\Gamma)) + \mathbb{P}(\sqrt{n}(\|W\|_2^2 + 1/n_1)^{1/2} > \eta/(2M')) \leq \rho$$

which is eventually smaller than $\rho$. Since $\eta, \rho$ were arbitrary we conclude that $B(W; \hat{f}_0) \to 0$ in probability so that $\hat{\tau}_{\text{WLS}(W)} \to 0$ in probability.

Consider case (b). Let $\bar{\beta}_0 = (\alpha_0, \beta_0^T)^T$. Then $\tilde{\beta} - \bar{\beta}_0 = \hat{P}^{-1}\hat{H}$ so by the above $\tilde{\beta} - \bar{\beta}_0 = O_p(n^{-1/2})$. Noting that $\hat{\tau}_{\text{WLS}(W)} - \text{SATT} = \frac{1}{n_1}\sum_{i \in \mathcal{T}_1} \epsilon_i + (\bar{\beta}_0 - \tilde{\beta})^T(\frac{1}{n_1}\sum_{i \in \mathcal{T}_1} \tilde{X}_i)$ completes the proof. $\square$

PROOF OF THM. 16. Writing $W_i' = (1 - T_i)W_i - T_i/n_1$, $K' = K + \lambda I_{\mathcal{T}_0}$,

we have that $\text{SKOM}(\mathcal{W}, \mathcal{K}, \lambda)$ is given by

$$\min_{W \in \mathcal{W}} \quad \sup_{\substack{\sum_{i,j=1}^{\infty} \alpha_i \alpha_j \mathcal{K}(x_i, x_j) \leq 1, \\ g \in \mathcal{G}, \sum_{i=1}^{\infty} \alpha_i^2 \mathcal{K}(x_i, x_i) < \infty, \\ \sum_{i=1}^{\infty} \alpha_i g'(x_i) = 0 \; \forall g' \in \mathcal{G}}} \quad \sum_{i=1}^{n} W_i' \left( g(X_i) + \sum_{j=1}^{\infty} \alpha_i \mathcal{K}(x_j, X_i) \right) + \lambda \|W\|_2^2$$

$$= \min_{W \in \mathcal{W}} \left\{ \begin{array}{ll} \infty & \exists g \in \mathcal{G} : \sum_{i=1}^{n} W_i' g(X_i) \neq 0 \\ W'^T K' W' & \forall g \in \mathcal{G} : \sum_{i=1}^{n} W_i' g(X_i) = 0 \end{array} \right. = \min_{\substack{W \in \mathcal{W}, \\ GW' = 0}} W'^T K' W'.$$

The result follows by writing $GW' = 0$ as $W' \in \text{null}(G) = \text{span}(N)$, which is in turn written as $W' = NU$ for a new variable $U \in \mathbb{R}^k$.  $\square$

PROOF OF THM. 23. Let $\Delta = \mathbb{E}\left[\sum_{i \in \mathcal{T}_1} W_i X_i - \sum_{i \in \mathcal{T}_0} W_i X_i\right]$ and similarly define $\Delta'$.

Suppose $\Delta = \alpha \Delta'$ for $\alpha \in [-1, 1]$. Then, for any $f(x) = \beta^T x + \beta_0$, $|\mathbb{E}[B(W; f)]| = |\beta^T \Delta| = |\alpha| |\beta^T \Delta'| \leq |\beta^T \Delta'| = |\mathbb{E}[B(W'; f)]|$.

Now, suppose $W$ is linearly EPBR relative to $W'$. Then, for any $\beta$, we have that if $\beta^T \Delta' = 0$ then $B(W'; f) = 0$ for $f(x) = \beta^T x$, which by linear EPBR implies that $B(W; f) = 0$, which means that $\beta^T \Delta = 0$. In other words, $\text{span}(\Delta')^\perp \subseteq \text{span}(\Delta)^\perp$. Therefore, $\text{span}(\Delta) \subseteq \text{span}(\Delta')$, which exactly means that $\exists \alpha \in \mathbb{R}$ such that $\Delta = \alpha \Delta'$. If $\Delta' = 0$ then we can choose $\alpha = 0$. If $\Delta' \neq 0$ then there exists $\beta$ such that $|\beta^T \Delta'| > 0$ and hence $|\alpha| = |\beta^T \Delta| / |\beta^T \Delta'| \leq 1$ by linear EPBR.  $\square$

PROOF OF THM. 24. Fixing any $\mu \in \mathbb{R}^d$, $\mu \neq 0$, we can affinely transform the data so that $X \mid T = 0$ is spherical at zero (has zero mean, unit covariance, and its distribution is unitarily invariant) and $X \mid T = 1$ is distributed the same as $\mu + \alpha X \mid T = 0$ for some $\alpha \in \mathbb{R}_+$. For any affinely invariant $W'$, we may assume this form for the data and any affinely invariant $W'$ is also unitarily invariant so that by spherical symmetry, $\mathbb{E}\left[\sum_{i=1}^{n} (-1)^{T_i+1} W_i X_i\right] \in \text{span}(\mu)$. Both $W$ and $W^{(0)}$ are affinely invariant.  $\square$

PROOF OF THM. 25. Fix $X$, $T$, $A \in \mathbb{R}^{d \times d}$, and $a \in \mathbb{R}^d$ with $A$ nonsingular. Let $\hat{\mu}$ and $\widehat{\Sigma}$ be defined as in Alg. 3 for $X$, $T$. Let $\hat{\mu}_A$ and $\widehat{\Sigma}_A$ be defined as in Alg. 3 for the transformed data $X_A = XA^T + \mathbf{1}_n a^T$, $T$. Then, $\hat{\mu}_A = A\hat{\mu} + a$ and $\widehat{\Sigma}_A = A\widehat{\Sigma}A^T$. The inner products produced by Alg. 3 on the transformed data are $((X_A)_i - \hat{\mu}_A)^T \widehat{\Sigma}_A^\dagger ((X_A)_j - \hat{\mu}_A) = (AX_i - A\hat{\mu})^T A^{-T} \widehat{\Sigma}^\dagger A^{-1} (AX_j - A\hat{\mu}) = (X_i - \hat{\mu})^T \widehat{\Sigma}^\dagger (X_j - \hat{\mu})$, which are the inner products produced by Alg. 3 on the untransformed data.  $\square$

PROOF OF THM. 26. If $X_{1:n}$ is affinely independent then it can be affinely mapped to the $(n-1)$-dimensional simplex, composed of $0$ and the first $n-1$ unit vectors. If $W$ is affinely invariant then it takes the same value on all affinely independent $X_{1:n}$ as it does on the $(n-1)$-dimensional simplex. □

PROOF OF THM. 27. The proof is the same as that of Thm. 23. □

PROOF OF THM. 28. The proof is the same as that of Thm. 24: without loss of generality we may assume that the embedded control data is spherical (ellipsoidal with zero mean and identity covariance operator and therefore spherically symmetric under unitary transformations) and that treated data is distributed like scaling and shifting the control data. □

PROOF OF THM. 29. The proof is the same as that of Thm. 25. □

PROOF OF THM. 30. Let $X_{1:n}$ and $X'_{1:n}$ each be a list of $n$ distinct elements of $\mathcal{X}$. Since $\mathcal{K}$ is strictly positive definite, both $\{\mathcal{K}(X_i, \cdot) : i = 1, \ldots, n\}$ and $\{\mathcal{K}(X'_i, \cdot) : i = 1, \ldots, n\}$ are linearly independent sets of vectors. Therefore, there exists a bounded non-singular $A \in \mathcal{F} \otimes \mathcal{F}$ such that $AX_i = X'_i$. Since $W$ is $\mathcal{K}$-affinely invariant, it is the same for $X_{1:n}$ and $X'_{1:n}$. □